\documentclass[10pt,twocolumn,letterpaper]{article}

\usepackage{cvpr}
\usepackage{times}
\usepackage{epsfig}
\usepackage{graphicx}
\usepackage{amsmath}
\usepackage{amssymb}
\usepackage{pifont}
\usepackage{nidanfloat}
\usepackage[utf8]{inputenc}
\usepackage{multirow}
\usepackage{float}
\usepackage{comment}
\usepackage[toc,page]{appendix}

\DeclareFixedFont{\ttb}{T1}{txtt}{bx}{n}{8} % for bold
\DeclareFixedFont{\ttm}{T1}{txtt}{m}{n}{8}  % for normal

\usepackage{color}

\definecolor{deepblue}{rgb}{0,0,0.5}
\definecolor{deepred}{rgb}{0.8,0,0}
\definecolor{deepgreen}{rgb}{0,0.5,0}
\definecolor{grey1}{rgb}{0.5,0.5,0.5}
\usepackage{listings}

\newcommand\pythonstyle{\lstset{
		language=Python,
		basicstyle=\small\ttm,
		otherkeywords={self},             % Add keywords here
		keywordstyle=\small\ttb\color{deepblue},
		emph={},          % Custom highlighting
		emphstyle=\small\color{deepred},    % Custom highlighting style
		stringstyle=\small\color{deepgreen},
		commentstyle=\small\color{grey1}\ttm,
		frame=tb,                         % Any extra options here
		showstringspaces=false,            % 
		breaklines=true,
		tabsize=2
}}
\lstnewenvironment{python}[1][] {
	\pythonstyle
	\lstset{#1}
}{}

\newcommand\pythoninline[1]{{\pythonstyle\lstinline!#1!}}

\makeatletter
\newcommand{\printfnsymbol}[1]{%
  \textsuperscript{\@fnsymbol{#1}}%
}
\makeatother
\cvprfinalcopy % *** Uncomment this line for the final submission

\def\cvprPaperID{5792} % *** Enter the CVPR Paper ID here

\ifcvprfinal\fi
\begin{document}

%%%%%%%%% TITLE
%In Defense of Pre-Trained ImageNet Architectures for Real-Time Semantic Segmentation of Road-driving Images
\title{In Defense of Pre-trained ImageNet Architectures\\
       for Real-time Semantic Segmentation 
       of Road-driving Images}

\author{Marin Oršić\thanks{equal contribution} \qquad Ivan Krešo\printfnsymbol{1} \qquad Petra Bevandić \qquad Siniša Šegvić\\
Faculty of Electrical Engineering and Computing\\
University of Zagreb, Croatia\\
{\tt\small name.surname@fer.hr}
}
% deadline 7:59 AM Saturday, Greenwich Mean Time (GMT)

\maketitle
\thispagestyle{empty}

%%%%%%%%% ABSTRACT
\begin{abstract}
Recent success of semantic segmentation approaches
on demanding road driving datasets 
has spurred interest in many 
related application fields.
Many of these applications involve 
real-time prediction on mobile platforms 
such as cars, drones and various kinds of robots.
Real-time setup is challenging due to 
extraordinary computational complexity involved.
Many previous works address the challenge
with custom lightweight architectures
which decrease computational complexity
by reducing depth, width and layer capacity 
with respect to general purpose architectures.
We propose an alternative approach
which achieves a significantly better performance
across a wide range of computing budgets.
First, we rely on a light-weight 
general purpose architecture
as the main recognition engine.
Then, we leverage light-weight upsampling
with lateral connections as
the most cost-effective solution
to restore the prediction resolution.
Finally, we propose to enlarge the receptive field 
by fusing shared features at multiple resolutions 
in a novel fashion.
Experiments on several road driving datasets show 
a substantial advantage of the proposed approach,
either with ImageNet pre-trained parameters 
or when we learn from scratch.
Our Cityscapes test submission entitled SwiftNetRN-18
delivers 75.5\% MIoU and achieves 39.9\,Hz 
on 1024$\times$2048 images on GTX1080Ti.
% Source code is available at \texttt{https://github.com/orsic/swiftnet}
\end{abstract}

\begin{figure}[htb]
  \centering
  \includegraphics[width=\columnwidth]{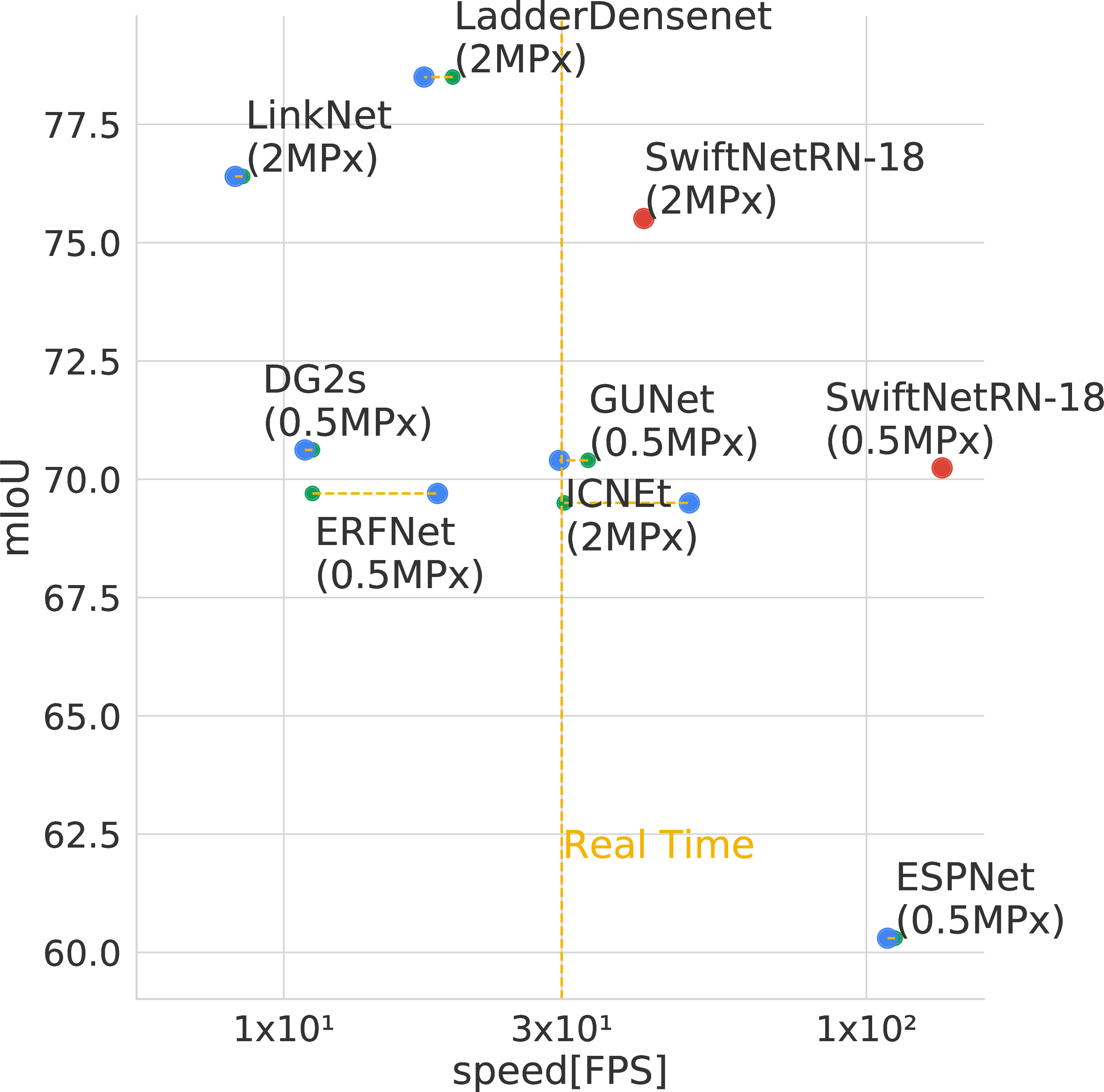}
  \caption{Speed-accuracy trade-off for different
    semantic semantic segmentation approaches
    on Cityscapes test on GTX1080Ti
    (except for DG2s which reports only validation performance).
    Red dots represent our method. 
    Other methods are displayed in green, 
    whereas blue dots show estimated frame rates 
    of the corresponding methods on our GPU
    (please refer to subsection \ref{ssec:comp} for details).
    Our submissions achieve the best accuracy and the best speed
    among all approaches aiming at real-time operation.
  }
  \label{fig:speed_miou}
\end{figure}

%%%%%%%%% BODY TEXT
\section{Introduction}

% Semantic segmentation, dense prediction
Semantic segmentation is an important 
dense prediction task in which 
the inference targets posterior distribution 
over a known set of classes in each image pixel 
\cite{farabet13pami,long15cvpr,chen17corr}.
Currently, the best results are achieved
with deep fully convolutional models which
require extraordinary computational resources.
Many important applications such as 
autonomous navigation or driver assistance
require inference on very large images
in order to cover a wide field of view
and perceive pedestrians at distances of over 200m.
At the same time, these applications 
require a very low latency in order 
to be able to bring real-time decisions.
The resulting requirements 
intensify computational strain
and make real-time implementations
a challenging research objective.

Many real-time semantic segmentation approaches
\cite{romera2018ieee,zhao2017icnet,mehta18eccv,siam18cvpr}
address this goal by introducing 
custom lightweight architectures
which are not suited 
for large-scale visual recognition. 
Most of these approaches 
initialize training from scratch,
and thus miss a huge regularization opportunity 
offered by knowledge transfer \cite{oquab14cvpr} 
from larger and more diverse 
recognition datasets \cite{russakovsky15ijcv}.
Consequently, these methods incur 
a comparatively large overfitting risk.
Some approaches alleviate this shortcoming
by pre-training on ImageNet \cite{romera2018ieee}.
However, our experiments suggest that 
the resulting benefits tend to be smaller 
than in architectures which are designed 
for competitive ImageNet performance.

A simple model for semantic segmentation 
starts with a fully convolutional encoder
which gradually decreases the resolution
and increases the number of feature maps
of the resulting representation.
Instead of performing global pooling
(as we would do in image-wide classification)
one can proceed by attaching a pixel-wise loss
to obtain the predictions \cite{farabet13pami}. 
The resulting model would lead to
a very fast evaluation on modern hardware,
however its accuracy would be rather low 
due to the following problems. 
Firstly, small objects (e.g.\ distant traffic signs) 
would not be recognized due to 
low resolution of pixel-wise predictions,
which is usually 32 times smaller than the input image. 
Secondly, the receptive field of such models 
would not be large enough 
to classify pixels at large objects
(e.g.\ nearby buses or trucks).
These problems can be alleviated 
with various techniques
such as dilated convolutions \cite{yu16iclr},
learned upsampling \cite{long15cvpr}, 
lateral connections \cite{rasmus15nips,ronneberger15miccai,lin17cvpr,kreso17cvrsuad}
and resolution pyramids \cite{farabet13pami}.
However, not all of these techniques 
are equally suited for real-time operation.

% U uvodu:
In this paper, we argue that a competitive blend
of efficiency and prediction accuracy can be achieved
by models based on lightweight 
ImageNet-grade classification architectures
\cite{he16cvpr,sandler18arxiv}.
Additionally, we propose a novel approach to increase 
the receptive field of deep model predictions,
based on a resolution pyramid 
with shared parameters \cite{farabet13pami}.
The proposed approach incurs a very modest 
increase of the model capacity
and is therefore especially suited
for datasets with large objects
and few annotated images.
Finally, we show that the resolution of the predictions
can be efficiently and accurately upsampled
by a lightweight decoder with lateral connections 
\cite{rasmus15nips,ronneberger15miccai}.
We show that the resulting semantic segmentation models
can be evaluated under various computing budgets,
and that they are feasible even on embedded GPU platforms.
We present experiments both with ImageNet pre-training
and learning from scratch 
on different road driving datasets.
Our experiments achieve state-of-the art 
semantic segmentation accuracy
among all existing approaches 
aiming at real-time execution.

%------------------------------------------------------------------------
\section{Related Work}

% % semseg
%     \cite{zhao17iccv}
%     \cite{lin17cvpr}
%     \cite{chen18eccv}
% % real time semseg
%     \cite{romera2018ieee}
%     \cite{zhao2017icnet}
%     \cite{paszke16arxiv}
%     \cite{chaurasia17vcip}
%     \cite{mazzini18bmvc}
%     \cite{mehta18eccv}
%     \cite{siam18cvpr}
%     \cite{vallurupalli18cvpr}
%     \cite{nekrasov18bmvc}
%     \cite{briot18cvpr}
% % D307
%     \cite{kreso16gcpr}
%     \cite{kreso17cvrsuad}
%     \cite{krapac16gcpr}
% % models
%     \cite{lin17cvpr}
%     \cite{ronneberger15miccai}
%     \cite{sandler18arxiv}
%     \cite{huang17cvpr}
%     \cite{zoph18cvpr}
%     \cite{singh18cvpr}
%     \cite{kim18eccv}
%     \cite{zhang18cvpr}
%     \cite{huang17corr}
% % usual suspects
%     \cite{krizhevsky12nips}
%     \cite{he16corr}
%     \cite{he16cvpr}
%     \cite{ioffe15icml}
%     \cite{kingma14corr}
% % datasets
%     \cite{russakovsky15ijcv}
%     \cite{cordts15cvpr}
%     \cite{brotsow08prl}
% % other
%     \cite{loshchilov16arxiv}

% Related work
% ===========
% Semantička segmentacija: 
%  * baseline - imagenet model, otpiliti global pool,
%    podići rezoluciju bileninearnom intepolacijom
%  * problemi: mali objekti (niska rezolucija), 
%    veliki objekti (nedovoljno receptivno polje)
%  * gustu predikciju možemo ostvariti npr. mehaničkim progušćivanjem,
%    naučenim naduzorkovnajem (FCN) i encoder-decoder arhitekturom (najbolje)
%  * veliko receptivno polje možemo ostvariti npr. dilatiranom konvolucijom
%    (problem: memorija, speed), SPP (sklon overfittingu?), 
%     piramidalna organizacija s dijeljenim parametrima (memorija?)

As described in the introduction, 
semantic segmentation models 
have to face two major problems:
restoring the input resolution
and increasing the receptive field.
The simplest way to restore input resolution
is to avoid downsampling.
%In order to increase the receptive field 
%of convolutions in the deep layers,
This is usually achieved by replacing 
stride-2 poolings with non-strided poolings,
and doubling the dilation factor
in subsequent convolutions
\cite{chen16cvpr,yu17cvpr}.
However, this approach increases
the resolution of deep latent representations,
which implies a large computational complexity.
Furthermore, dilated convolutions 
incur significant additional slow-down
due to necessity to rearrange the image data
before and after calling optimized implementations.

Another way to achieve dense image prediction 
relies on trained upsampling \cite{long15cvpr},
which results in an encoder-decoder architecture.
The idea is to perform recognition 
on subsampled latent representation
to reduce complexity and then
to restore the resolution by upsampling 
the representation (or the predictions). 
This setup can be naturally augmented 
by introducing lateral connections \cite{rasmus15nips,ronneberger15miccai,lin17cvpr,kreso17cvrsuad}
to blend semantically rich deep layers
with spatially rich shallow layers.
%This setup can easily accomodate
%lightweight ImageNet-grade models 
%in the downsampling path. 
The upsampling path has to be as lean as possible 
(to achieve efficiency and prevent overfitting)
but no leaner (to avoid underfitting).
It turns out that the sweet spot is 
computationally inexpensive, 
which makes this setup especially well-suited 
for real-time operation.
% approaches aiming at 

Early approaches to enlarge 
the receptive field of logit activations
were based on dilated convolutions
\cite{yu16iclr,chen16cvpr,yu17cvpr}.
% A more involved approach is known as 
% spatial pyramid pooling (SPP) \cite{he15ieee,lazebnik06cvpr};
% it concatenates spatial poolings of various sizes 
% before calculating the logits.
% The convolutional variant of SPP 
% extracts features of different 
% spatial pooling grids \cite{zhao17iccv}.
A more involved approach is known as
spatial pyramid pooling (SPP) \cite{he15ieee}.
SPP averages features over aligned grids
with different granularities \cite{lazebnik06cvpr}.
We use a convolutional adaptation of that idea,
in which the feature pyramid is upsampled
to the original resolution \cite{zhao17iccv}
and then concatenated with the input features.
Thus, subsequent convolutions obtain access
to broad spatial pools %of input features,
and that increases their receptive field.
The combination of dilated convolutions and SPP
is known as \` a trous SPP, or ASPP for short \cite{chen17corr}.
However, SPP and ASPP may hurt generalization
due to large capacity.
In this paper, we study resolution pyramids 
as an alternative way to increase the receptive field 
and at the same time promote scale invariance \cite{singh18cvpr,kreso16gcpr,lin17cvpr2,chen16cvpr}.
Most previous pyramidal approaches to semantic segmentation
\cite{farabet13pami,zhao2017icnet,mazzini18bmvc}
fuse only the deepest representations 
extracted at different resolutions.
Different from them, we combine representations
from different abstraction levels 
before joining the upsampling path within the decoder.
This results in a better gradient flow
throughout the pyramidal representation
which is advantageous when the objects are large
and the training data is scarce.

% Efikasne arhitekture
% * grouped convolutions
% * depthwise separable convolution
% * factorized convolutions
% * piramide
% ** originalno su se koristile za scale invarance
% \cite{farabet,kreso,snip,mazzini}
% ** mi se nismo usrećili u tom smjeru, 
%    pretpostavljamo da je to 
%    zbog velikog kapaciteta modela
% ** međutim, mi piramidu koristimo za proširenje receptivnog polja
% * prva grupa pristupa se temelje na arhitekturama 
%   koje nisu prikladne za ImageNet
%   kako bi ostvarili kompaktniji model
%  ** opisati predstavnici ...
% * druga grupa pristupa temelji se na lakim
%   klasifikacijskim arhitekturama za ImageNet
%   ** opsiati ImageNet arhitekture: 
%      ResNet18, DenseNet-121, MobileNet, Xception, ShuffleNet?
%      (možda tablica s brojem parametara, slojeva, gflopsa, top-1)
%   ** začudo, ovakav pristup nije istražen u postojećoj literaturi 
% * postoji još i mogućnost pruninga, ali nije popularna

% Najsličniji radovi i razlika u odnosu na njih:
% * LinkNet, ICNet, brzi RefineNet, DG2s, GUNet

Efficient recognition architectures 
leverage optimized building blocks 
which aim at reducing computational load 
while preserving accuracy.
Grouped convolutions reduce 
the number of floating point operations 
and the number of parameters 
by enclosing the information flow 
within smaller groups of feature maps. 
Various methods have been proposed
to discover prominent inter-group connections.
ShuffleNet \cite{zhang18cvpr} uses channel shuffling 
to pass information across convolutional groups. CondenseNet \cite{huang17corr} incorporates 
a training strategy which locates
important connections in grouped convolutions 
and prunes those which are redundant. 
Neural architecture search \cite{zoph18cvpr}
is an approach that leverages reinforcement learning
to jointly learn the model architecture 
and the corresponding parameters.
The resulting architectures achieve 
competitive ImageNet performance 
when the computational budget is restricted.
Depthwise separable convolutions
\cite{sifre14corr,wang17iccv}  
decrease computational complexity
by splitting a regular convolution in two. 
Firstly, a $k \times k$ convolution 
is separably applied to each input channel.
This can be viewed as a group convolution
where the number of groups corresponds 
to the number of channels C.
In other words, there are C kernels k$\times$k$\times$1. 
Secondly, a 1$\times$1 convolution is applied 
to propagate inter-channel information.
Replacing standard convolutions 
with depthwise separable convolutions 
lowers the number of parameters 
and increases the inference speed 
at the cost of some drop in performance \cite{howard17corr}.
Strong regularization effect
of depthwise separable convolutions 
can be relaxed by inverted residual blocks  \cite{sandler18arxiv} 
which lead to compact residual models 
suitable for mobile applications.
%Residual models alleviate the problem of vanishing %gradients and achieve high accuracy %\cite{he16cvpr,he16corr}.

Most semantic segmentation approaches 
aiming at real-time operation
refrain from using encoders
designed for competitive ImageNet performance.
ICNet \cite{zhao2017icnet} proposes a custom encoder 
which processes an image pyramid with shared parameters
and fuses multi-scale representations 
before entering the decoder which restores the resolution.
ERFNet \cite{romera2018ieee} redefines a residual block as a composition of a $3 \times 1$ followed by a $1 \times 3$ convolution, which yields a 33\% reduction in parameters. 
Vallurupalli et al. \cite{vallurupalli18cvpr} 
propose the DG2s approach
as an efficient ERFNet variant with the 
following modifications in residual blocks: 
i) depthwise separable convolutions, and
ii) channel shuffling operation before pointwise  convolutions.
ESPNet \cite{mehta18eccv} factorizes convolutions
in a similar manner and refrains 
from shared parameters across the image pyramid
in order to produce a fast and compact architecture.

Our method is most related to 
semantic segmentation approaches 
which use lightweight encoders trained on ImageNet 
and benefit from such initialization.
Similar to our work, Nekrasov et al.\ \cite{nekrasov18bmvc} 
rely on MobileNet V2 \cite{sandler18arxiv} and 
NASNet \cite{zoph18cvpr} encoders 
and feature a thick decoder with lateral connections.
This is similar to our single scale model,
however, our decoder has much less capacity,
which allows us to report a half 
of their number of floating point operations
without sacrificing recognition accuracy 
on road driving datasets.
%Additionally we propose a novel pyramidal fusion approach
%which considerably improves our CamVid results.
%Unfortunately we cannot compare the accuracy
%since they do not report results on road driving datasets.
%which has more 
%optimize the RefineNet architecture \cite{lin17cvpr2} 
%by replacing $3 \times 3$ with $1 \times 1$ convolutions in the decoder 
%and also omitting RCU blocks.
LinkNet \cite{chaurasia17vcip} uses a small ResNet-18 backbone 
and a lightweight decoder to achieve satisfying performance-speed ratio. 
Our single scale model is similar to LinkNet, 
however we omit convolutional layers at full resolution 
in order to substantially reduce memory imprint
and greatly increase the processing speed.
Mazzini et al. \cite{mazzini18bmvc} 
use a dilated encoder initialized 
from the DRN-D-22 model \cite{yu17cvpr},
and a decoder with one lateral connection.
They also learn nonlinear upsampling 
to improve accuracy at object boundaries.
Instead of using dilated convolutions,
our decoder upsamples predictions
by exclusively relying on lateral connections,
which results in a 4-fold speed-up.
Additionally, we succeed to leverage
full resolution images during training
which achieves an improvement of 
5 percentage points on Cityscapes test.

\section{The proposed segmentation method}

Our method assumes the following requirements. 
The model should be based on an 
ImageNet pre-trained encoder 
in order to benefit from regularization induced by transfer learning. 
The decoder should restore the resolution of encoded features 
in order for the predictions to retain detail. 
The upsampling procedure must be as simple as possible 
in order to maintain real-time processing speed. 
Gradient flow should be promoted throughout the network 
to support training from random initialization 
in an unusual event that ImageNet pre-training 
turns out not to be useful.

\subsection{Basic building blocks}

The proposed segmentation method 
is conceived around three basic building blocks
which we briefly describe in the following paragraphs.
These building blocks are going to be used in 
our two models which we propose
in subsequent subsections.

\paragraph{Recognition encoder}
We advocate the usage of compact ImageNet pre-trained 
model as segmentation encoders. 
We propose to use ResNet-18 \cite{he16cvpr} 
and MobileNet V2 \cite{sandler18arxiv} 
for a number of reasons. 
These models are a good fit for fine tuning
due to pre-trained parameters being publicly available.
They are also suitable for training from scratch
due to moderate depth and residual structure.
Finally, these models are compatible
with real-time operation 
due to small operation footprint.
Computationally, ResNet-18 is around six times 
more complex than MobileNet V2.
However, MobileNet V2 uses 
depthwise separable convolutions
which are not directly supported
in GPU firmware (the cuDNN library).
Therefore, MobileNet V2 tends to be slower 
than ResNet-18 in most experimental setups.
Note that the same issue disqualifies
usage of the DenseNet architecture \cite{huang17cvpr},
since it requires efficient convolution 
over a non-contiguous tensor,
which is still not supported in cuDNN.

\paragraph{Upsampling decoder}
The recognition encoder transforms the input image 
into semantically rich visual features.
These features must have a coarse spatial resolution
in order to save memory and processing time.
The purpose of the decoder 
is to upsample these features
to the input resolution. 
We advocate a simple decoder organized
as a sequence of upsampling modules
with lateral connections
\cite{rasmus15nips,ronneberger15miccai}.
The proposed ladder-style 
upsampling modules 
have two inputs: 
the low resolution features 
(which should be upsampled), 
and the lateral features 
from an earlier layer of the encoder.
The low resolution features 
are first upsampled with bilinear interpolation 
to the same resolution as the lateral features. 
%This usually means upsampling them by a factor of 2. 
Upsampled input features 
and lateral encoder features 
are then mixed with elementwise summation
and finally blended 
with a 3$\times$3 convolution. 
%The final output of the upsampling module 
%is  of the summed features. 

\begin{figure}[h]
  \centering
  \includegraphics[width=\columnwidth]{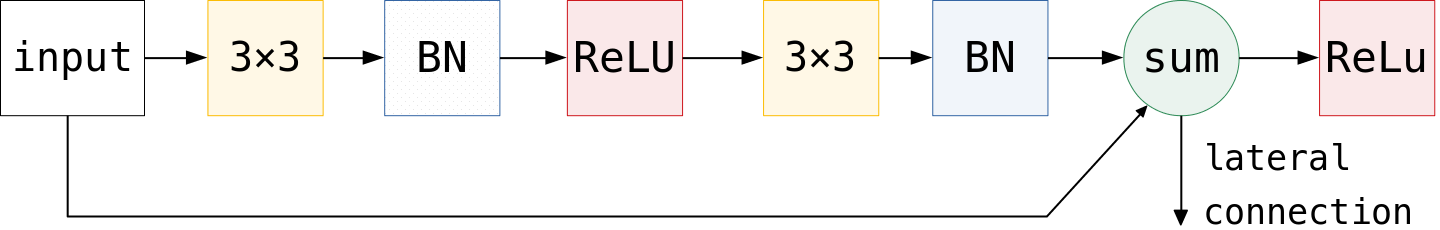}
  \caption{Structural diagram 
    of the last residual unit
    within a convolutional block 
    operating on common spatial resolution. 
    We do not use pre-activation     \cite{he16eccv}
    since we could not find
    a pre-trained parameterization 
    for ResNet-18.
    The lateral connection is taken
    from the output of the elementwise sum 
    after the last residual block.
    The output of the ReLU node is forwarded 
    to the next residual block.
  }
  \label{fig:resblock}
\end{figure}

We propose to route the lateral features
from the output of the elementwise sum 
within the last residual block
at the corresponding level of subsampling, 
as shown in Figure~\ref{fig:resblock}.
Note that routing the lateral features
from the output of the subsequent ReLU
leads to a considerable drop 
in validation accuracy.
Replacing the standard 3$\times$3 convolution 
with either a 1$\times$1 convolution, 
or a depthwise separable convolution
also decreases the validation accuracy. 
%TODO: try additional 3x3 layer here
%This suggests that the upsampling
%path requires a certain level of capacity.

\paragraph{Module for increasing the receptive field} 
As discussed before, there are two viable possibilities 
for increasing the receptive field 
while maintaining real-time speed:
i) spatial pyramid pooling, and ii) pyramid fusion.
The SPP block gathers the features
produced by the encoder at several pooling levels
and produces a representation with a varying level of detail.
We demonstrate the use of SPP in our single scale model. 
Our SPP block is a simplified version 
of the pyramid pooling module from PSPNet \cite{zhao17iccv}.
The pyramid fusion produces genuine 
multi-scale representations 
which need to be carefully fused within the decoder
in order to avoid overfitting to unsuitable level of detail.
We propose a pyramid pooling approach which blends 
representations at different levels of abstraction
and thus enlarges the receptive field 
without sacrificing spatial resolution.

\subsection{Single scale model}
\label{ssec:single}
\begin{figure}[htb]
  \centering
  \includegraphics[width=\columnwidth]{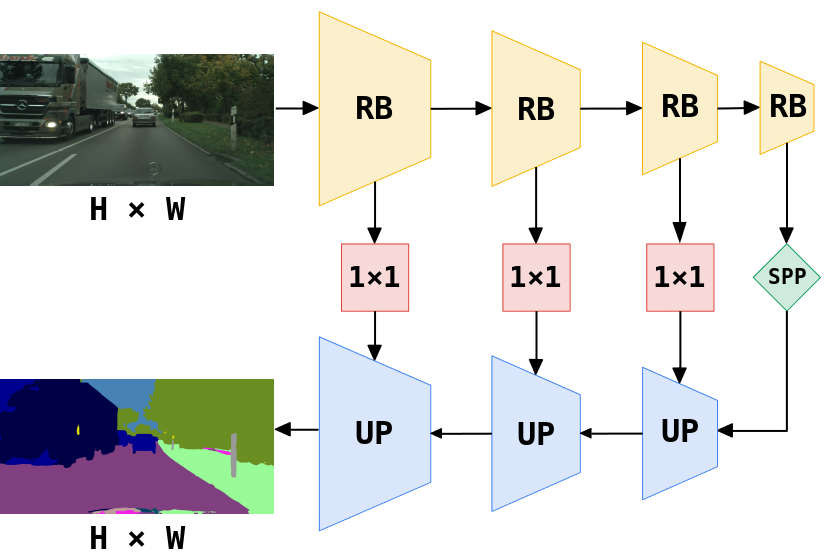}
  \caption{Structural diagram of the proposed single scale model. 
    Yellow trapezoids designate convolution groups 
    within the encoder which may be pre-trained on ImageNet. 
    The green diamond designates the spatial pyramid pooling layer, 
    the red squares designate bottleneck layers, 
    and blue trapezoids designate lightweight upsampling modules. 
    Logits are upsampled to original image resolution 
    with bilinear interpolation.
}
  \label{fig:baseline}
\end{figure}

The proposed single scale model transforms 
the input image into dense semantic predictions
throughout a downsampling recognition encoder
and upsampling decoder, as shown in Figure~\ref{fig:baseline}.
%The features produced by all considered encoders
%are 32 times subsampled with respect to the input image. 
%Resolution of these features should be enlarged in order to retrieve detail, %especially for small distant objects. 
%The receptive field should also be enlarged 
%to be able to correctly classify pixels of large objects.
Yellow trapezoids designate convolution groups,
that is, parts of the encoder which operate
on the same spatial resolution.
All considered encoders consist of four such convolution groups.
The first convolution group produces features
at the H/4$\times$W/4 resolution,
while each following group increases the subsampling by the factor of 2.
Thus the features at the far end of the encoder are H/32$\times$W/32.
These features are fed into 
the spatial pyramid pooling layer
(designated by a green diamond)
in order to increase 
the model receptive field.
The resulting tensor is 
finally routed to the decoder 
whose upsampling modules are shown in blue.

Note that decoder and encoder are asymmetric:
the encoder has many convolutions 
per convolution group while decoder has 
only one convolution per upsampling module.
Furthermore, the dimensionality of encoder features
increases along the downsampling path,
while the dimensionality of the decoder features is constant.
Therefore, lateral connections have to adjust dimensionality 
with 1$\times$1 convolutions designated with red squares.
Upsampling modules operate in three steps: 
i) the low resolution representation is bilinearly upsampled, 
ii) the obtained representations is summed with the lateral connection, 
iii) the summation is blended using a 3$\times$3 convolution.

\subsection{Interleaved pyramid fusion model}
\label{ssec:pyr}

While using a compact encoder is beneficial for fast inference, 
this also results in a decreased receptive field 
and a smaller capacity compared to general purpose 
convolutional models for visual recognition. 
To counteract these drawbacks, we propose
to exploit image pyramids 
to enlarge the receptive field of the model 
and reduce the model capacity requirements.

\begin{figure}[htb]
  \centering
  \includegraphics[width=\columnwidth]{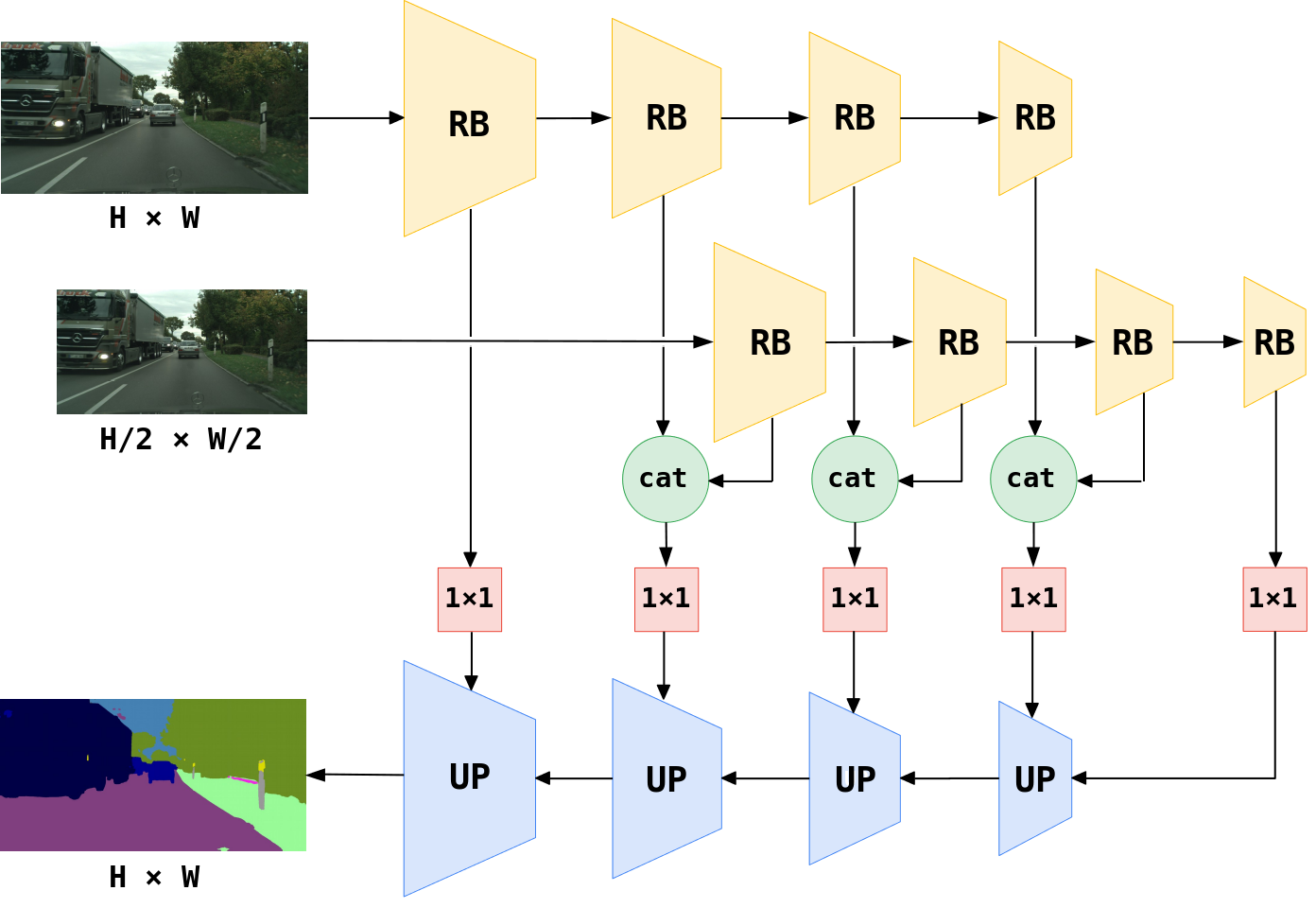}
  \caption{%
    Structural diagram of the proposed model 
    with interleaved pyramidal fusion. 
    Encoder parameters (yellow) are shared across all pyramid levels and 
    may be pre-trained on Imagenet.
    Features of the same resolutions are concatenated (green circles), 
    fed into a 1$\times$1 bottleneck convolution (red squares) 
    and blended within the decoder (blue).
  }
  \label{fig:cascade}
\end{figure}

The proposed model is shown in Figure~\ref{fig:cascade}. 
Two encoder instances (yellow) are applied to the input image
at different levels of the resolution pyramid.
This results in increased receptive field 
of the activations which sense
the lowest resolution of the image pyramid.
Furthermore, shared parameters enable 
recognition of objects of different sizes
with the common set of parameters,
which may relax the demand for model capacity.
In order to enforce lateral connections
and improve the gradient flow throughout the encoder,
we concatenate the feature tensors 
from neighbouring levels of different encoders
(we can do that since they have equal spatial resolution).
This concatenation is designated with green circles.
After concatenation, interleaved encoder features
are projected onto the decoder feature space
by 1$\times$1 convolutions designated with red squares. 
The decoder (blue) operates in the same manner
as in the single-scale model, 
however now we have an additional upsampling module
for each additional level of the image pyramid.
%We refrain from sharing the feature projection parameters 
%since that decreased the validation performance. 
%The representations from the common feature space 
%are concatenated and fed to the upsampling modules 
%same as in the baseline model.

%------------------------------------------------------------------------
\section{Experiments}

We conduct semantic segmentation experiments on two datasets: 
Cityscapes \cite{cordts15cvpr} and CamVid \cite{brotsow08prl}. 
We report mIoU accuracy, computational complexity 
and the execution speed of the trained models. 
The speed measurements are performed on a desktop GPU (GTX 1080Ti)
and on an embedded System on a chip module (Jetson TX2).
We also present ablation and validation experiments 
which provide a more detailed insight 
into the impact of various design choices.
Please note that additional experiments
can be found in the supplement.

\subsection{Training details}

We train all our models with the Adam \cite{kingma14corr} optimizer
with the learning rate set to $4\cdot 10^{-4}$. 
We decay the learning rate with cosine annealing \cite{loshchilov16arxiv} 
to the minimum value of $1\cdot 10^{-6}$ in the last epoch
(we do not perform any warm restarts). 
The weight decay is set to $1\cdot 10^{-4}$. 
In experiments with ImageNet pre-training, 
we update pre-trained parameters 
with 4 times smaller learning rate 
and apply 4 times smaller weight decay. 
We train on jittered square crops with batch size 12. 
The jittering consists of random horizontal flipping, 
and scaling with random factors between 0.5 and 2.
We use 768$\times$768 crops for full Cityscapes resolution,
and 448$\times$448 crops for half Cityscapes resolution and CamVid.
We train for 200 epochs on Cityscapes and 400 epochs on CamVid. 
We train for additional 200 epochs in experiments 
without ImageNet pre-training.

\subsection{Measuring the computational complexity}
\label{ssec:comp}

We express the computational complexity 
of the trained models with two metrics: 
i) billions of floating point operations (GFLOP), and 
ii) number of processed frames per second (FPS). 
The GFLOP metric provides the number of
fused multiply-add operations 
required to evaluate the model.
Such platform-agnostic measure of the computational complexity
is suitable for CPUs where all multiplications require 
roughly equal processing time.
Unfortunately, the GFLOP metric poorly corresponds
with the actual processing time on GPU platforms,
since efficient implementations are available
only for a small subset of all building blocks 
used to express current deep models.
Consequently, both metrics are required for
a complete description of algorithm suitability for real-time operation.
%This slows down or effectively disqualifies
%several exciting architectures \cite{huang17cvpr,sandler18arxiv},
%although the support will probably get better in the future.

The FPS metric directly corresponds to the processing time
on a particular hardware platform.
Such metric does not necessarily correlate across platforms,
although rough estimations can be done, as we show below.
We simulate real-time applications
by setting batch size to 1.
We measure the time elapsed between 
transferring the input data to the GPU, 
and receiving the semantic predictions into RAM
as shown in the code snippet shown in Figure~\ref{fig:mtpt}.

\begin{figure}[htb]
    \centering
    \begin{python}
device = torch.device('cuda')
model.eval()
model.to(device)
with torch.no_grad():
  input = model.prepare_data(batch).to(device)
  logits = model.forward(input)
  torch.cuda.synchronize()
  t0 = 1000 * perf_counter()
  for _ in range(n):
    input = model.prepare_data(batch).to(device)
      logits = model.forward(input)
      _, pred = logits.max(1)
      out = pred.data.byte().cpu()
      torch.cuda.synchronize()
  t1 = 1000 * perf_counter()
fps = (1000 * n) / (t1 - t0)
\end{python}
    \caption{Measurement of the processing time under PyTorch.}
    \label{fig:mtpt}
\end{figure}

We conduct all measurements
on a single GTX1080Ti with 
CUDA 10.0, CUDNN 7.3 and PyTorch 1.0rc1.
We exclude the batch normalization layers \cite{ioffe15icml} 
from measurements since in real-time applications 
they would be fused with preceding convolutional layers. 
We report mean FPS values over 1000 forward passes. 
Results are shown in Table~\ref{tab:fps_gflops}. 
The column \emph{FPS norm} provides a rough estimate 
on how would other methods perform on our hardware. 
The scaling factors are: 1.0 for GTX1080Ti, 
0.61 for TitanX Maxwell, 1.03 for TitanX Pascal, and 1.12 for Titan XP.
These scaling factors were calculated using publicly 
available benchmarks: 
\texttt{goo.gl/N6ukTz},
\texttt{goo.gl/BaopYQ}.
The column \emph{GFLOPs@1MPx} shows 
an estimated number of FLOPs 
for an input image of 1MPx,
as a resolution-agnostic metric
of computational complexity.

\begin{table*}[b]
  \footnotesize
  \begin{center}
    \begin{tabular}{l|cccccccccc}
		model                                     &subset& mIoU          & FPS   & FPS norm & GPU            & resolution & GFLOPs & GFLOPs@1Mpx & \# params    \\
		\hline
		\hline
		D*   \cite{vallurupalli18cvpr}            & val  & 68.4          & -     & -     & TitanX M                   & 1024x512   &  5.8   & 11.6  & 0.5M  \\
		DG2s \cite{vallurupalli18cvpr}            & val  & 70.6          & -     & -     & TitanX M                   & 1024x512   &  19.0  & 38    & 1.2M  \\
		Ladder DenseNet\dag \cite{kreso17cvrsuad} & val  & 72.8          & 31.0  & 30.1  & TitanX                     & 1024x512   & -      & -     & 9.8M   \\
		ICNet \cite{zhao2017icnet}                & test & 69.5          & 30.3  & 49.7  & TitanX M                   & 2048x1024  &  -     & -     & -     \\
		ESPNet \cite{mehta18eccv}                 & test & 60.3          & 112   & 108.7 & TitanX                     & 1024x512   &  -     & -     & 0.4M  \\
		ERFNet \cite{romera2018ieee}              & test & 68.0          & 11.2  & 18.4  & TitanX M                   & 1024x512   & 27.7   & 55.4  & 20M   \\
		\hline
		GUNet\dag \cite{mazzini18bmvc}            & test & 70.4          & 37.3  & 33.3  & TitanXP                    & 1024x512   &  -     & -     & -     \\
		ERFNet\dag \cite{romera2018ieee}          & test & 69.7          & 11.2  & 18.4  & TitanX M                   & 1024x512   & 27.7   & 55.4  & 20M   \\
		%ERFNet\dag \cite{romera2018ieee}          & test & 69.7          & 11.2  & 18.4  & TitanX M                   & 1024x512   & 27.7   & 55.4  & 20M   \\
		%TODO ne valja fps norm stupac
		%LinkNet\dag \cite{chaurasia17vcip}       & val  & 76.4          & 8.5   & 8.5   & TitanX                     & 1920x1080  & 143.1  & 72.5  & 11.5M \\
		\hline
		SwiftNetRN-18                             & val  & 70.4          & 39.9  & 39.3  & \multirow{2}{*}{GTX 1080Ti}& 2048x1024  & 104.0  & 52.0  & 11.8M \\
		SwiftNetMN V2                             & val  & 69.4          & 27.7  & 27.7  &                            & 2048x1024  & 41.0   & 20.5  & 2.4M  \\
		\hline
		SwiftNetRN-18\dag                         & val  & 70.2          & 134.9 & 134.9 & \multirow{4}{*}{GTX 1080Ti}&  1024x512  & 26.0   & 52.0  & 11.8M \\
		SwiftNetRN-18 pyr\dag                     & val  & 74.4          & 34.0  & 34.0  &                            & 2048x1024  & 114.0  & 57.0  & 12.9M \\
		SwiftNetMN V2\dag                         & val  & 75.3          & 27.7  & 27.7  &                            & 2048x1024  & 41.0   & 20.5  & 2.4M  \\
		SwiftNetRN-18\dag                         & val  & 75.4          & 39.9  & 39.3  &                            & 2048x1024  & 104.0  & 52.0  & 11.8M \\
		\hline
		SwiftNetRN-18 pyr\dag                     & test  & 75.1         & 34.0  & 34.0  & \multirow{3}{*}{GTX 1080Ti}& 2048x1024  & 114.0  & 57.0  & 12.9M \\
		SwiftNetRN-18\dag                         & test  & \textbf{75.5}& 39.9  & 39.3  &                            & 2048x1024  & 104.0  & 52.0  & 11.8M \\
		SwiftNetRN-18 ens\dag                     & test  & \textbf{76.5}        & 18.4  & 18.4  &                    & 2048x1024  & 218.0  & 109.0 & 24.7M \\
	\end{tabular}
  \end{center}
  \caption{Results of semantic segmentation on Cityscapes. 
    We evaluate our best result on the online test benchmark
    and compare it with relevant previous work, where possible.
    We also report the computational complexity (GFLOP, FPS)
    GPU on which the inference was performed,
    and the image resolution on which 
    the training and inference were performed. 
    The column \emph{GFLOPs@1Mpx} shows  
    the GFLOPs metric when the input resolution is 1MPx. 
    The column FPS norm shows or estimates 
    the FPS metric on GTX 1080Ti.
    % SwiftNetRN-18 pyr denotes the pyramid fusion model
    The default SwiftNet configuration
    is the single scale model
    presented in \ref{ssec:single}.
    Label pyr denotes the pyramid fusion model
    presented in \ref{ssec:pyr}.
    % SwiftNetRN-18 pyr denotes the ensemble of single scale
    % and pyramid model.
    % All other SwiftNet experiments correspond to
    % single scale model presented in \ref{ssec:single}.
    % The symbol \dag designates ImageNet pre-training.}
    Label ens denotes the ensemble of
    the single scale model and the pyramid model.
    The symbol \dag{} designates ImageNet pre-training.}
  \label{tab:fps_gflops}
\end{table*}

\subsection{Cityscapes}

The Cityscapes dataset is a collection of high resolution images 
taken from the driver's perspective during daytime and fine weather. 
It consists of 2975 training, 500 validation, and 1525 test images 
with labels from 19 classes. 
It also provides 20000 coarsely labeled images 
which we do not use during our experiments.
Table~\ref{tab:fps_gflops} evaluates
the accuracy (class mIoU) and 
efficiency (GFLOP, FPS) of our methods
and compares them to other real-time methods. 
Our single scale method based on the ResNet-18 encoder
achieves 75.5\% test mIoU, and delivers 39.9\,FPS 
on full Cityscapes resolution (1024$\times$2048 pixels).
To the best of our knowledge, this result outperforms 
all other approaches aiming at real-time operation. 
The corresponding submission to the Cityscapes evaluation
server is entitled SwiftNetRN-18.
Table~\ref{tab:fps_gflops} also presents experiments 
in which our models are trained from scratch. 
The accuracy decreases 
for 5 mIoU percentage points (pp)
with respect to the corresponding experiments
with ImageNet pre-trained initialization. 
This shows that ImageNet pre-training 
represents an important ingredient 
for reaching highly accurate predictions. 
We notice that custom encoders like ERFNet \cite{romera2018ieee}
get less benefits from ImageNet pre-training: 
only 1.7\% pp as shown in Table~\ref{tab:fps_gflops}.
Figure~\ref{fig:cs_examples} presents examples 
of segmentations on Cityscapes val images. 
We show examples for both single scale and pyramid models.
We did not achieve measurable improvements
with the pyramid model 
over the single scale model
on the Cityscapes dataset.
%Table~\ref{tab:cs_scratch} shows 
%results when training the model 
%without ImageNet initializaion. 
%These results are comparable to other methods 
%which use random initialization, 
% however we experience a performance drop 
%compared to using ImageNet pre-training.

\begin{comment}
\begin{table}[ht]
%   \footnotesize
  \begin{center}
    \begin{tabular}{l|cc}
      backbone       &   model                        & mIoU     \\
      \hline\hline
      ResNet-18      &  \multirow{2}{*}{single scale} &  70.35   \\
      MobileNet V2   &                                &  69.44   \\
    \end{tabular}
  \end{center}
  \caption{Semantic segmentation accuracy 
    on Cityscapes val
    with random initialization.}
  \label{tab:cs_scratch}
\end{table}
\end{comment}

\subsection{CamVid}

The CamVid dataset contains 
701 densely annotated frames.
We use the usual split into 
367 train, 101 validation and 233 test images. 
We train on combined train and validation subsets
and evaluate semantic segmentation 
into 11 classes on the test subset. 
Table~\ref{tab:camvid} shows that we obtain 
an improvement of roughly 1.5 pp mIoU 
when using the pyramid model with 
pre-trained ResNet-18 and MobileNetV2 backbones.
Figure~\ref{fig:cv_examples} shows 
frames from the CamVid test subset
where the pyramid model performed better.

Table~\ref{tab:camvid} also shows 
that ImageNet pre-training
contributes more on CamVid than on Cityscapes
(7-9pp of mIoU performance).
This is not surprising since
CamVid has almost 20 times less training pixels.

A small size of the dataset 
poses a considerable challenge 
when training from scratch 
due to high overfitting risk. 
Table~\ref{tab:camvid} shows that the pyramid
model achieves better results 
than the single scale model. 
This supports our choice of sharing 
encoder parameters across pyramid levels.

\begin{table}[ht]
  %\footnotesize
  \begin{center}
    \begin{tabular}{l|ccc}
      backbone                        & model        &mIoU\dag& mIoU\\
      \hline\hline
    \multirow{2}{*}{ResNet-18}        & single scale & 72.58  & 63.33 \\
                                      & pyramid      & 73.86  & 65.70 \\
      \hline
    \multirow{2}{*}{MobileNet V2}     & single scale & 71.56  & 64.01 \\
                                      & pyramid      & 73.08  & 65.01 \\
    \end{tabular}
  \end{center}
  \caption{Semantic segmentation accuracy on CamVid test 
    using ImageNet pre-training (mIoU\dag) and training from scratch (mIoU).}
  \label{tab:camvid}
\end{table}

\subsection{Validation of the upsampling capacity}

The number of feature maps along the upsampling path 
is the most important design choice of the decoder.
We validate this hyper-parameter and report the results in Table~\ref{tab:upsample_ablation}.
The results show that the model accuracy saturates at 128 dimensions.
Consequently, we pick this value as a sensible 
speed-accuracy trade-off in all other experiments.

\begin{comment}
We investigate the susceptibility of ladder-style upsampling to overfitting. 
A single upsampling module is made of one 1$\times$1 convolution 
to project skip connection features to a feature 
space of same dimensionality as the upsampled features. 
The features are then summed up and blended using a $3 \times 3$ convolution. Thus, the most i
\end{comment}

\begin{table}[ht]
  %\footnotesize
  \begin{center}
    \begin{tabular}{l|ccc}
      model                               & upsampling features & mIoU     \\
  \hline\hline
      \multirow{4}{*}{SwiftNetRN-18}      & 64                  & 69.50   \\
                                          & 128                 & 70.35   \\
                                          & 192                 & 70.26   \\
                                          & 256                 & 70.63   \\
    \end{tabular}
  \end{center}
  \caption{Validation of the number of feature maps in the upsampling path.
    The models were trained on Cityscapes train subset at 512$\times$1024
    while the evaluation is performed on Cityscapes val. 
    All models use ImageNet initialization.
  }
  \label{tab:upsample_ablation}
\end{table}

\subsection{Ablation of lateral connections}

To demonstrate the importance of lateral connections 
between the encoder and the decoder, 
we train a single scale model without lateral connections. 
For this experiment, we discard the 1$\times$1 convolution layers 
located on the skip connections and abandon 
the elementwise summations in upsampling modules. 
Training such a model on full Cityscapes train images
causes the validation accuracy to drop from 75.35\% to 72.93\%.

% begin supplement

\subsection{Execution profile}

To obtain a better insight 
into the execution time of our models,
we report separate processing times 
and the GFLOP metrics
for the downsampling path 
(encoder and SPP), 
and the upsampling path (decoder). 
Table~\ref{tab:profile} shows the results
for the single scale model
and input resolution of 2048$\times$1024.
Note the striking discrepancy
of time and GFLOPs
for the two downsampling paths.
ResNet-18 is almost twice as fast
than MobileNet v2
despite requiring 6 times 
more multiplications.
Note also that our decoder
is twice as fast as 
the ResNet-18 encoder.

\begin{table}[ht]
    \begin{center}
    \begin{tabular}{l|ccccccc}
    model & dn time    & up time    & dn FLOPs & up FLOPs \\
    \hline
    \hline
    RN-18 & 15.0ms   & 8.1ms & 76.1B     & 30.9B     \\
    MN-V2 & 26.7ms   & 7.5ms & 12.1B     & 28.9B     \\
    \end{tabular}
    \end{center}
    \caption{Inference speed 
      along the downsampling (encoder and SPP) 
      and the upsampling (decoder) paths for the single scale model.
      The columns \textit{dn time} and \textit{up time} 
      display the execution times,
      while the columns \textit{dn FLOPs} and \textit{up FLOPs} 
      show the number of floating point operations for 2048$\times$1024 images.
    }
    \label{tab:profile}
\end{table}

\subsection{Size of the receptive field}
We estimate the effective receptive field 
of our models by considering the central pixel 
in each image $ \mathbf{X}$ of Cityscapes val.
The estimate is based on gradients $\frac{\partial y_i}{\partial \mathbf{X}}
$\cite{luo16nips}
where $\mathbf{y}$ are the logits for the central pixel
while $i$ is $argmax(\mathbf{y})$.
 Table~\ref{tab:erf} expresses the effective receptive fields 
as standard deviations of pixels with 
top 5\% gradient $\frac{\partial y_i}{\partial \mathbf{X}}$.
The results show that both SPP and interleaved pyramidal fusion substantially increase 
the receptive field.

\begin{table}[h]
    % \footnotesize
    \centering
    \begin{tabular}{l|cc}
     model            & ERF horizontal   & ERF vertical    \\
     \hline
     \hline
        RN-18 no SPP  & 84.47            & 84.78           \\
        RN-18 SPP     & 154.07           & \textbf{153.55} \\
        RN-18 pyramid & \textbf{185.22}  & 140.24          \\
    \end{tabular}
    \caption{Effective receptive fields (ERF)
    expressed as standard deviation
    of pixels with top 5\% image gradients with respect to the 
    dominant class of the central pixel,
    measured on Cityscapes val.}
    \label{tab:erf}
\end{table}

\subsection{Processing speed on Jetson TX2}

\begin{figure}[htb] 
  \centering
  \includegraphics[width=\columnwidth]{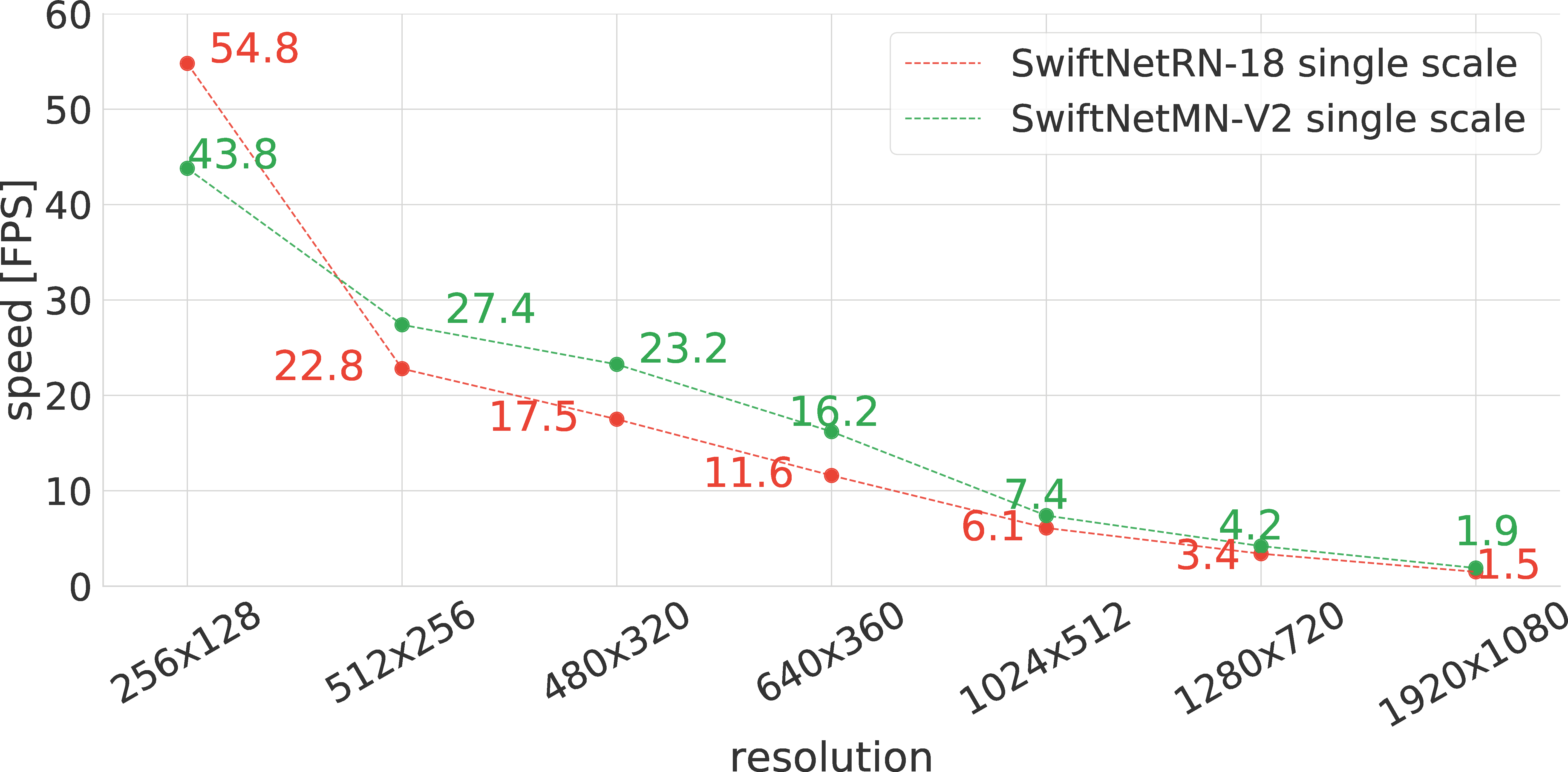}
  \caption{Processing speed in frames per second 
    of the proposed architecture
    on NVIDIA Jetson TX2 module 
    for two different backbones
    and various input resolutions.}
  \label{fig:tx2}
\end{figure}

We evaluate the proposed methods on NVIDIA Jetson TX2 module
under CUDA 9.0, CUDNN 7.1, and PyTorch v1.0rc1. 
Due to limited number of CUDA cores, 
all bilinear interpolations had to be replaced 
with nearest neighbour interpolation. 
Results are reported in Figure~\ref{fig:tx2}.
The MobileNet V2 backbone outperforms ResNet-18 
for 20-30\% on most resolutions
due to lower number of FLOPs.
However, ResNet-18 is faster 
on the lowest resolution.
Note that our implementations 
do not use TensorRT optimizations.

\newcommand{\sswt}{0.505\columnwidth}
\begin{figure*}[b]
    \centering
    \includegraphics[width=\sswt]{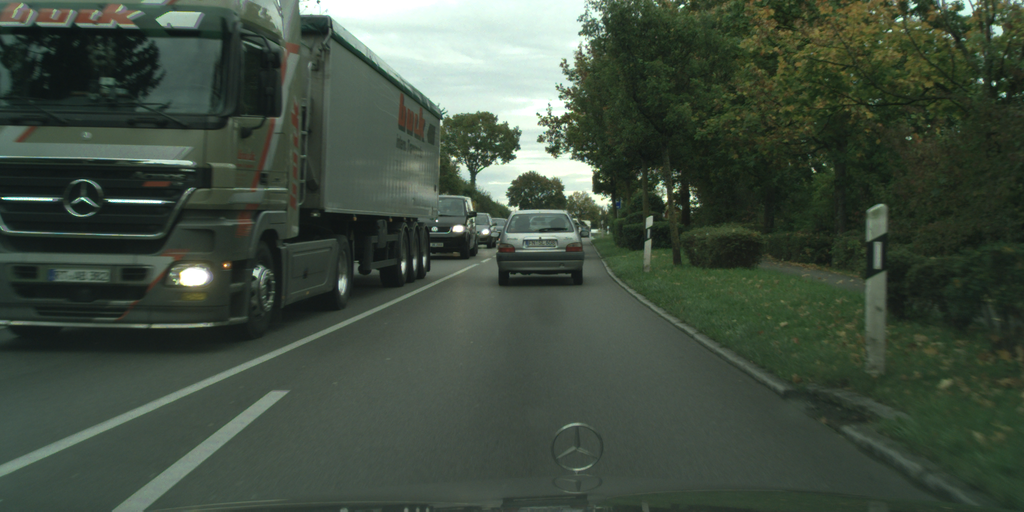}
    \hfill
    \includegraphics[width=\sswt]{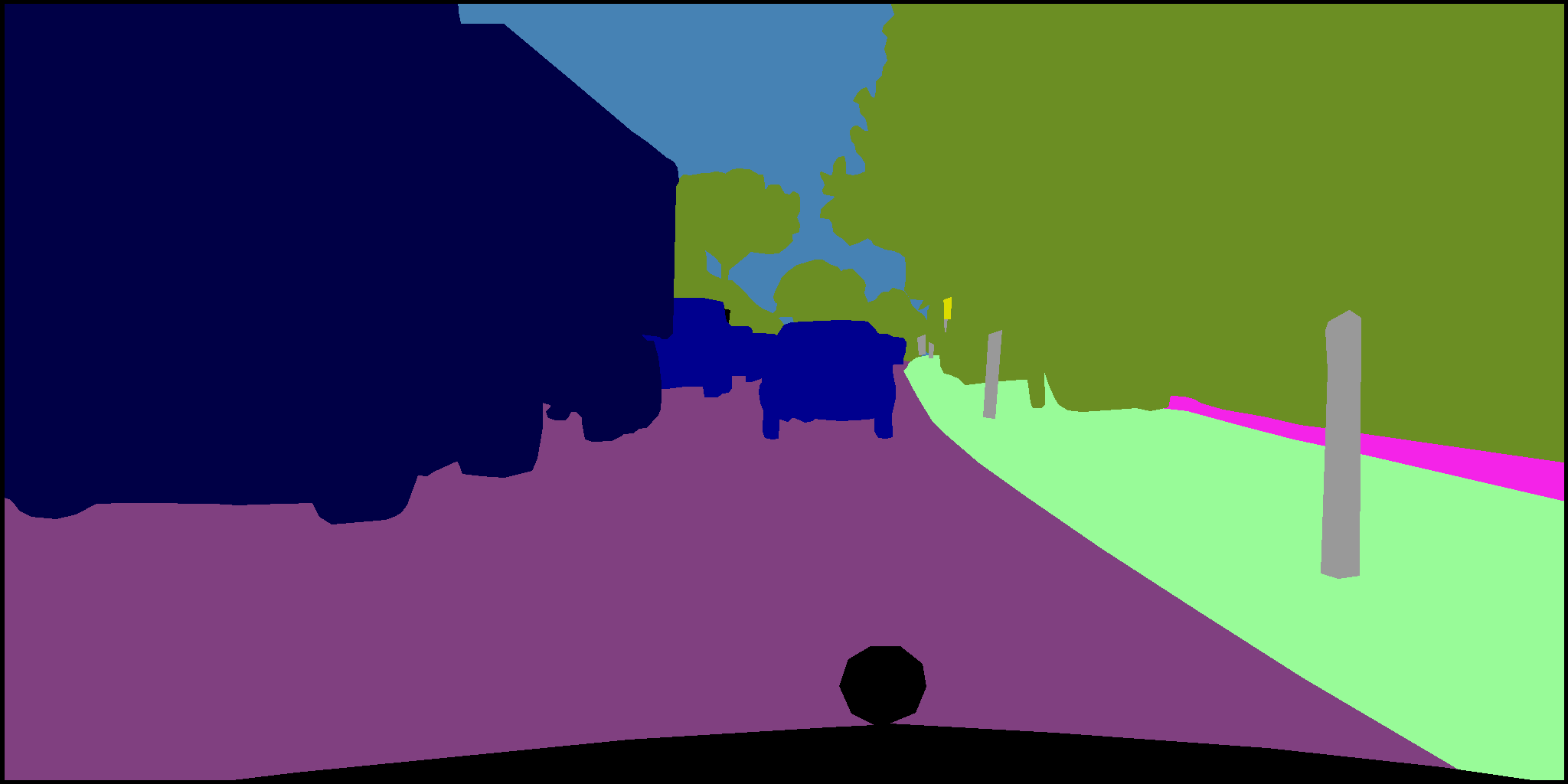}
    \hfill
    \includegraphics[width=\sswt]{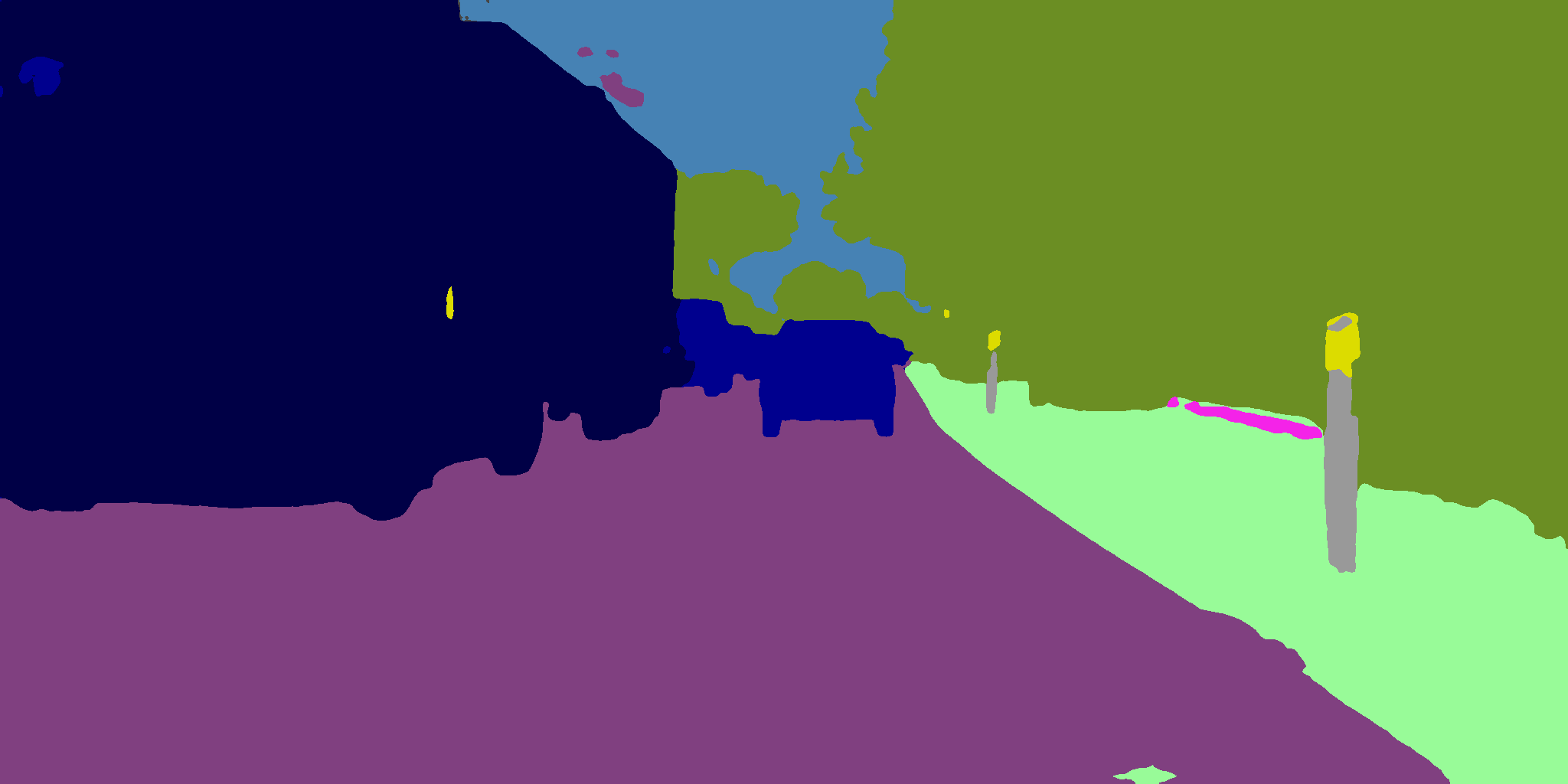}
    \hfill
    \includegraphics[width=\sswt]{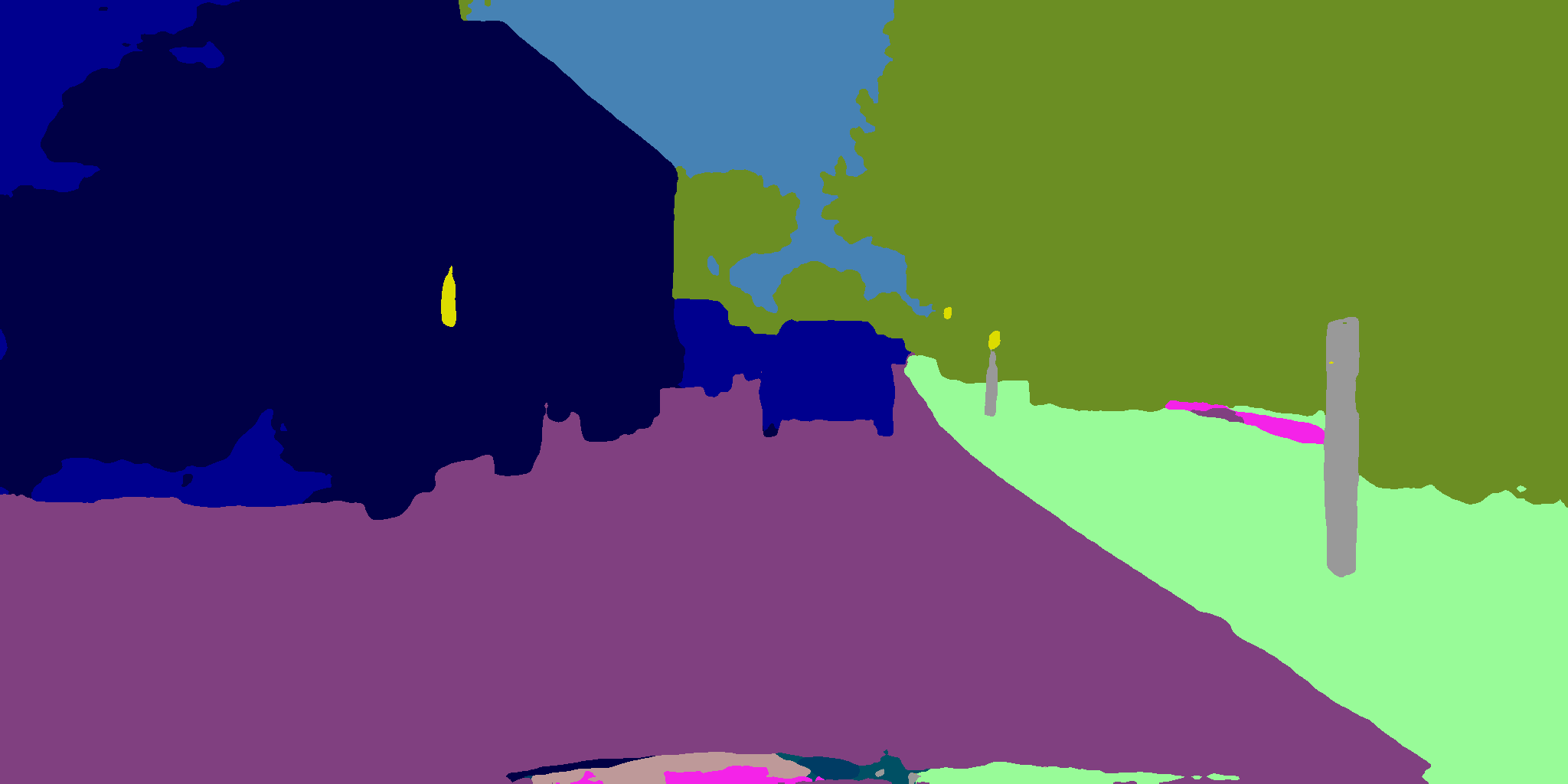}
    
    \includegraphics[width=\sswt]{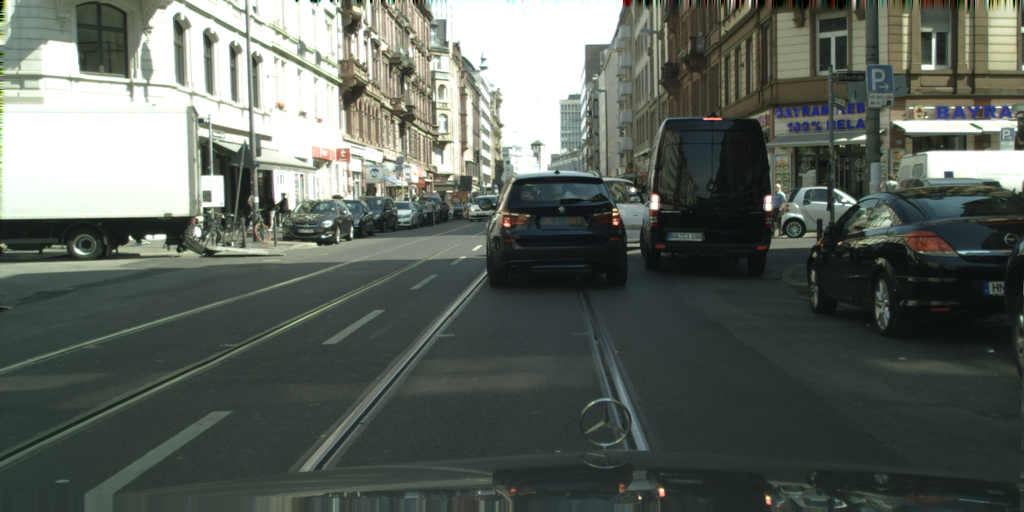}
    \hfill
    \includegraphics[width=\sswt]{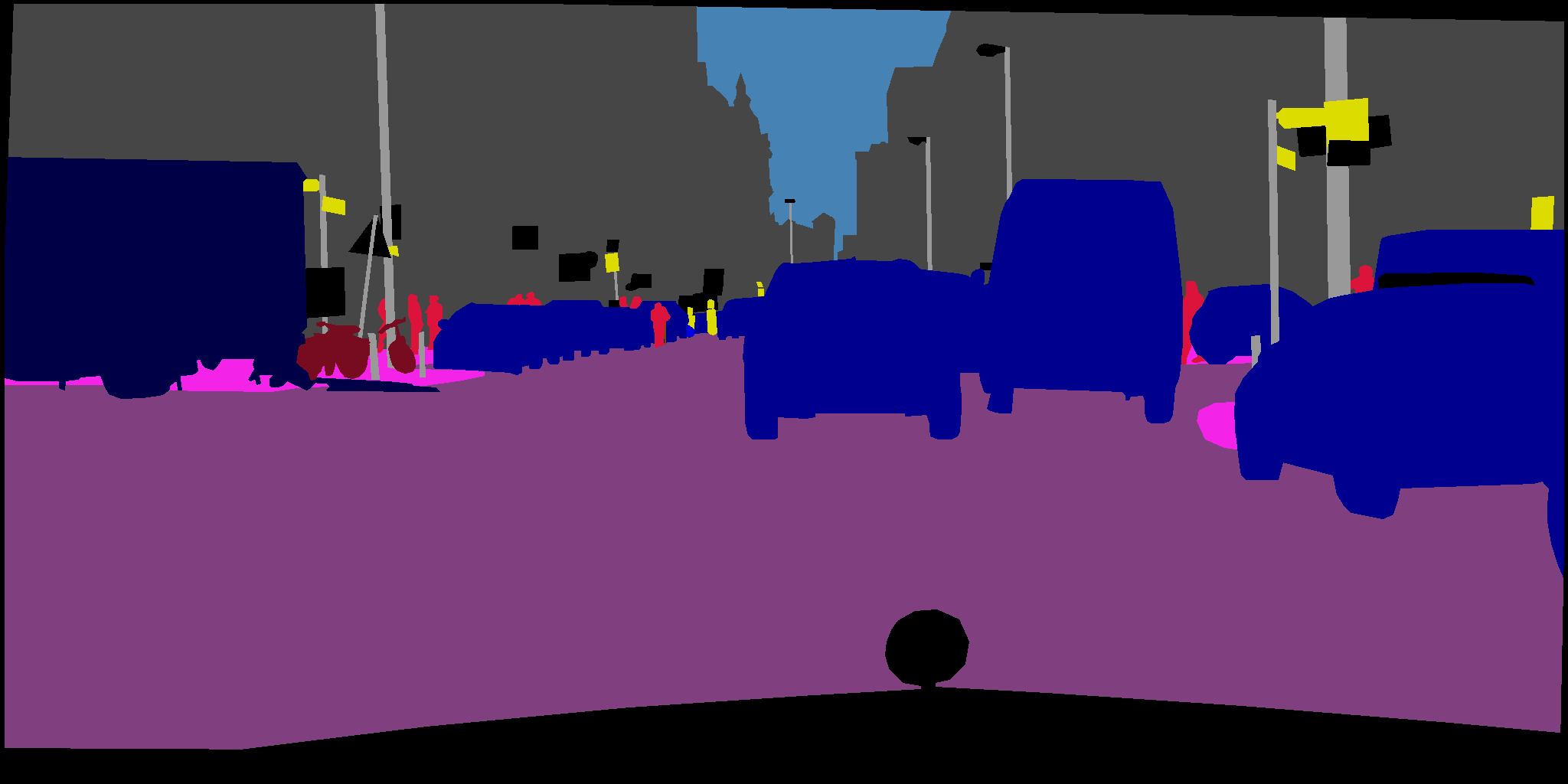}
    \hfill
    \includegraphics[width=\sswt]{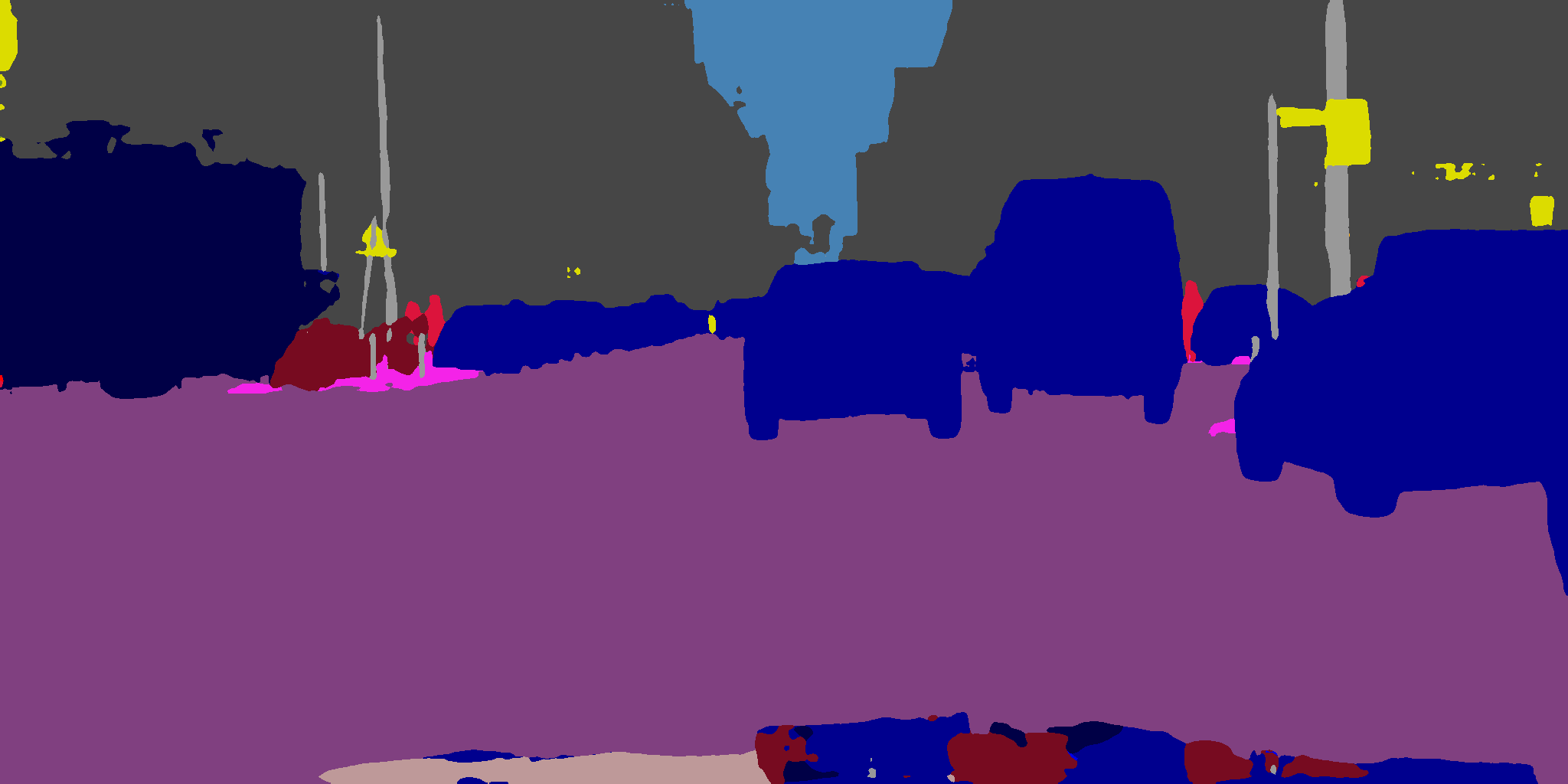}
    \hfill
    \includegraphics[width=\sswt]{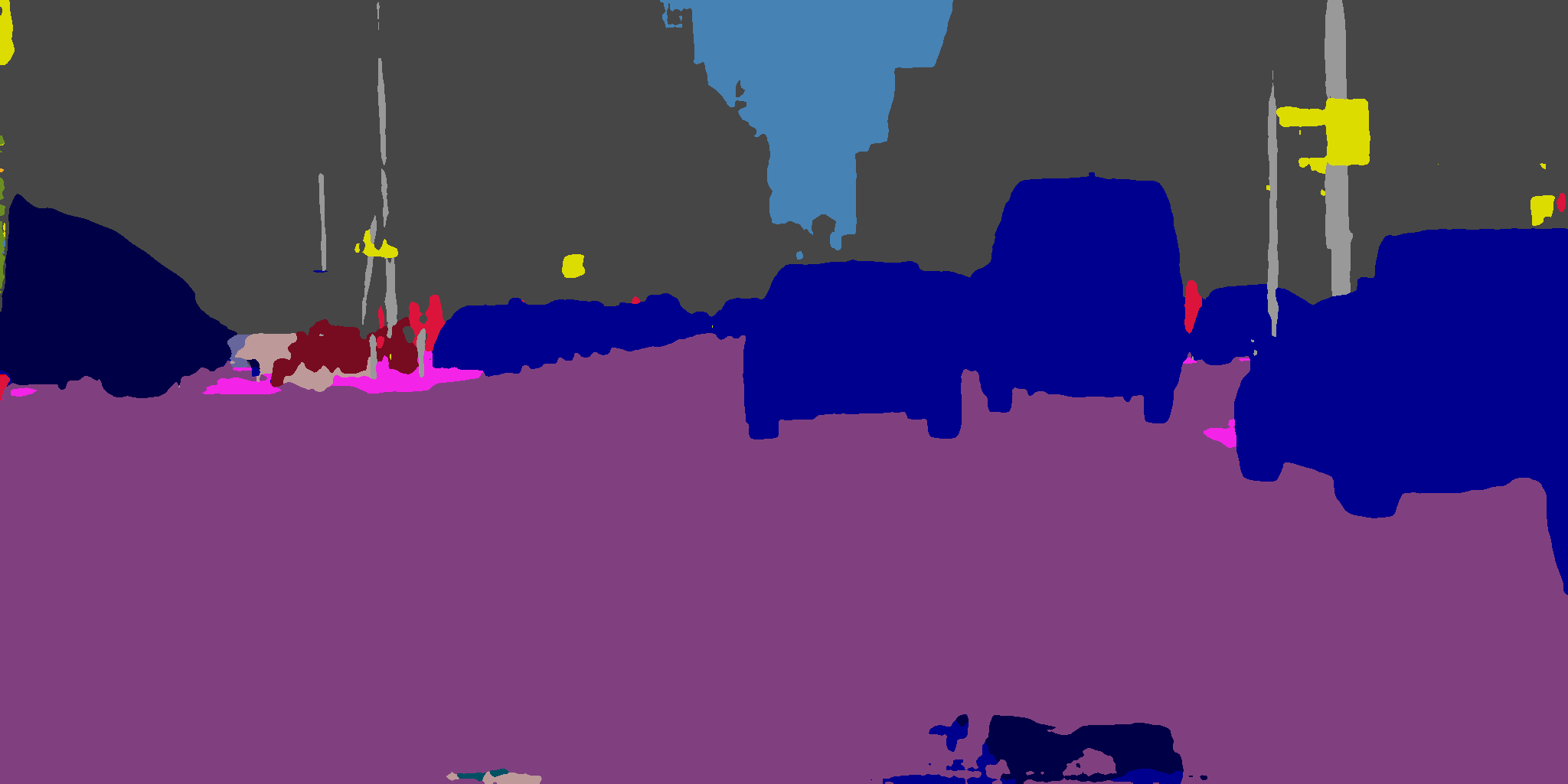}
    
    \includegraphics[width=\sswt]{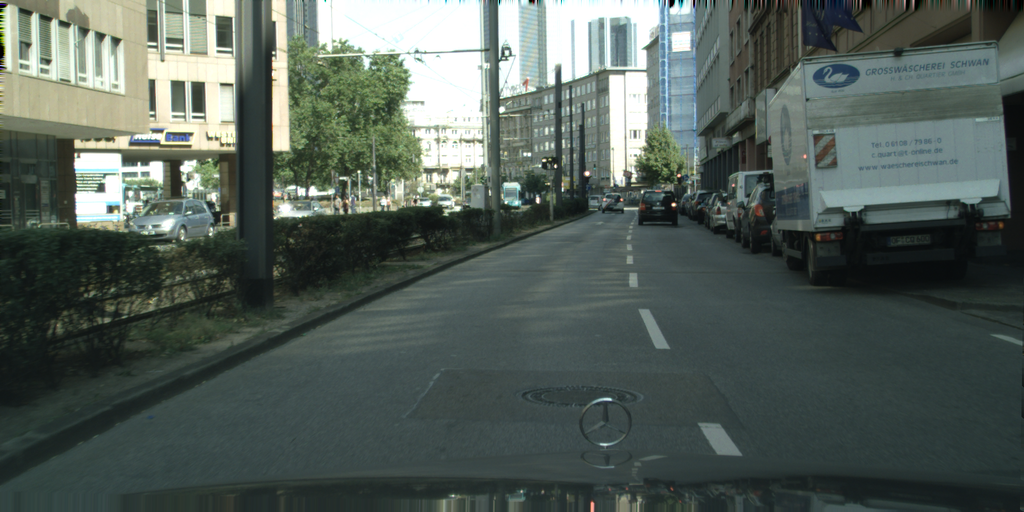}
    \hfill
    \includegraphics[width=\sswt]{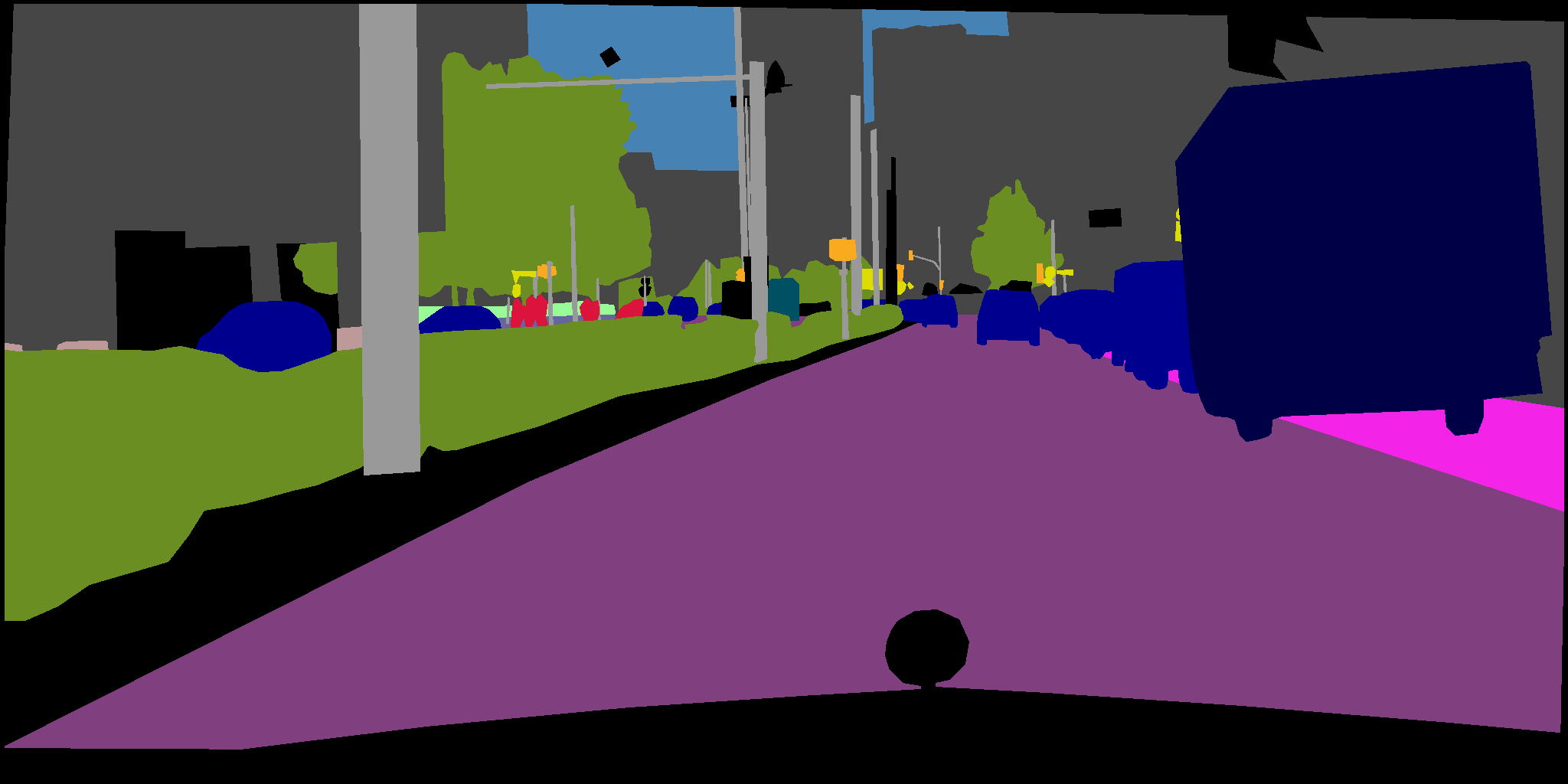}
    \hfill
    \includegraphics[width=\sswt]{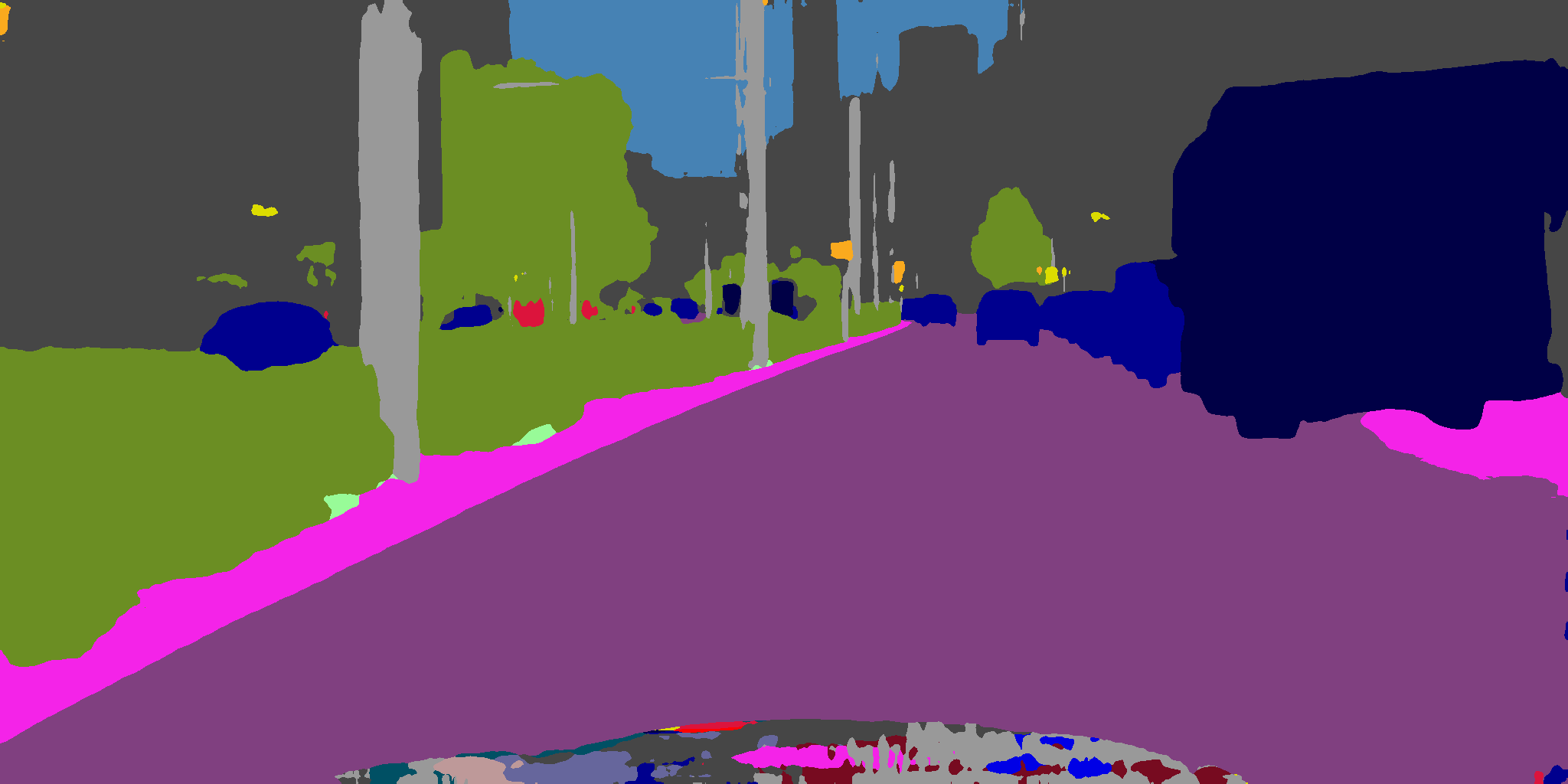}
    \hfill
    \includegraphics[width=\sswt]{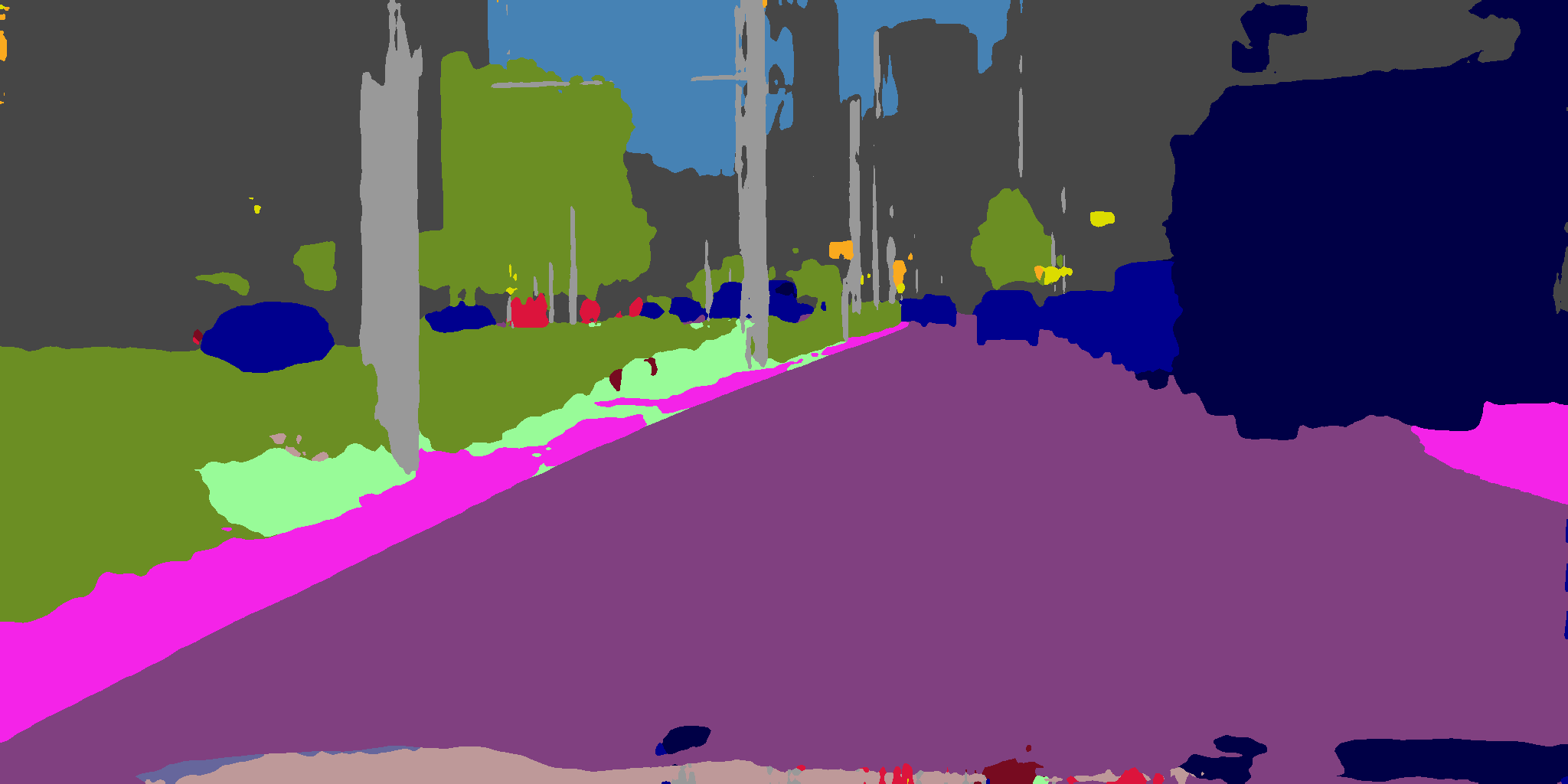}
    \caption{Semantic segmentation results on Cityscapes val. 
      The columns correspond to input image, 
      ground truth annotation, 
      the output of the pyramid model, 
      and the output of the single scale model.
      The most significant improvements
      occur on pixels of the class truck.
     }
    \label{fig:cs_examples}
\end{figure*}

\begin{figure*}[b]
    \centering
    \includegraphics[width=\sswt]{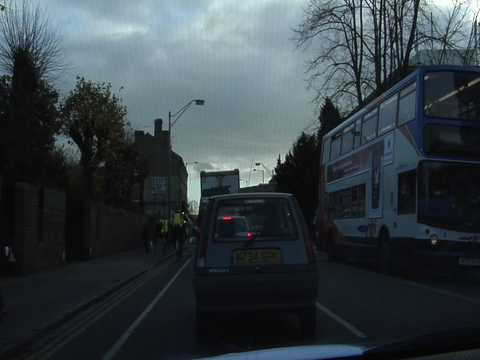}
    \hfill
    \includegraphics[width=\sswt]{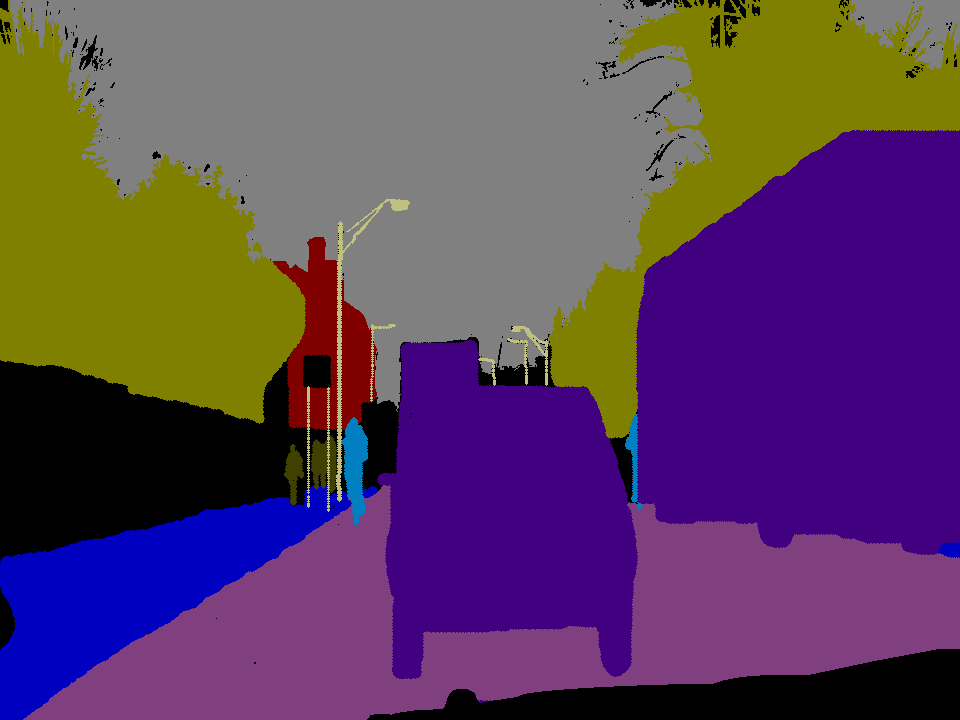}
    \hfill
    \includegraphics[width=\sswt]{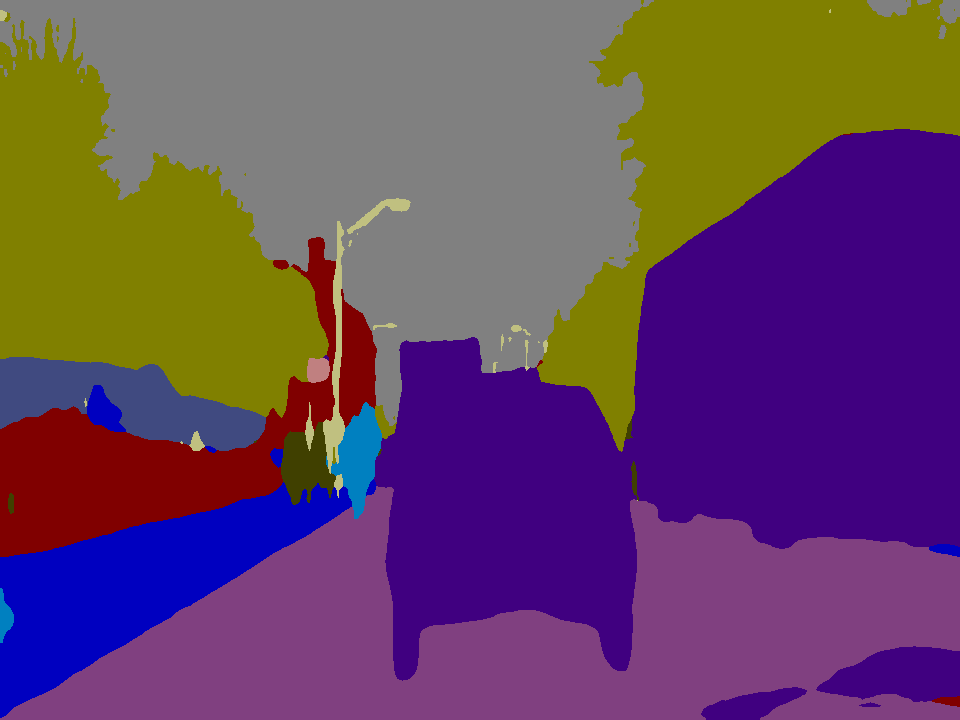}
    \hfill
    \includegraphics[width=\sswt]{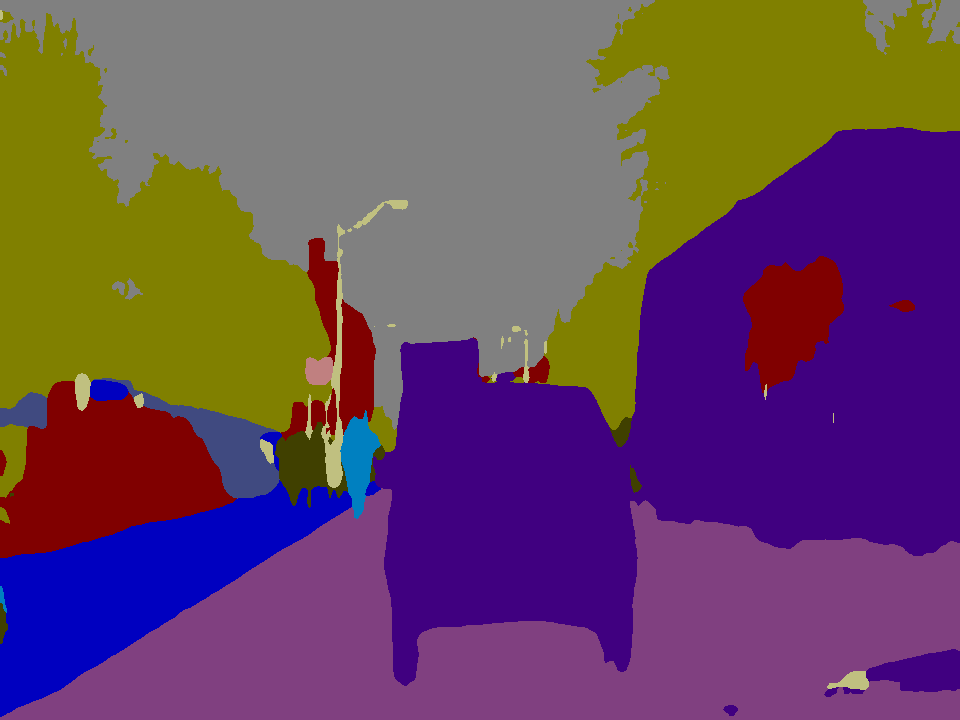}
    
    \includegraphics[width=\sswt]{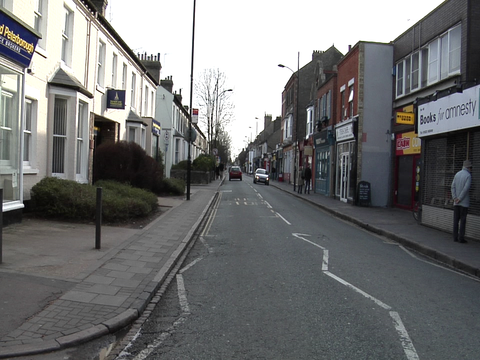}
    \hfill
    \includegraphics[width=\sswt]{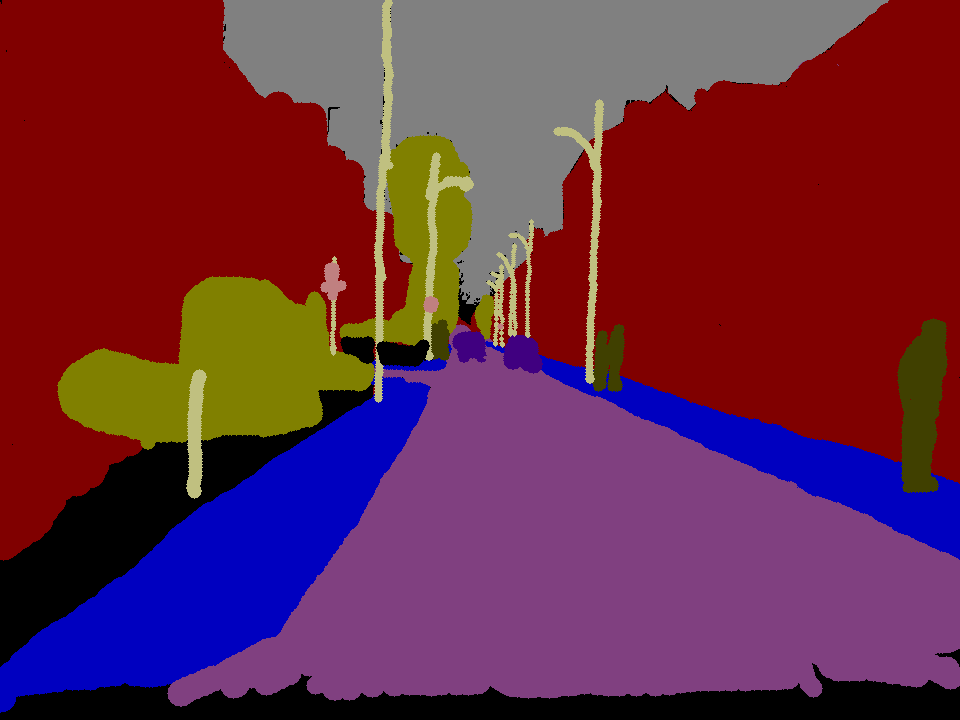}
    \hfill
    \includegraphics[width=\sswt]{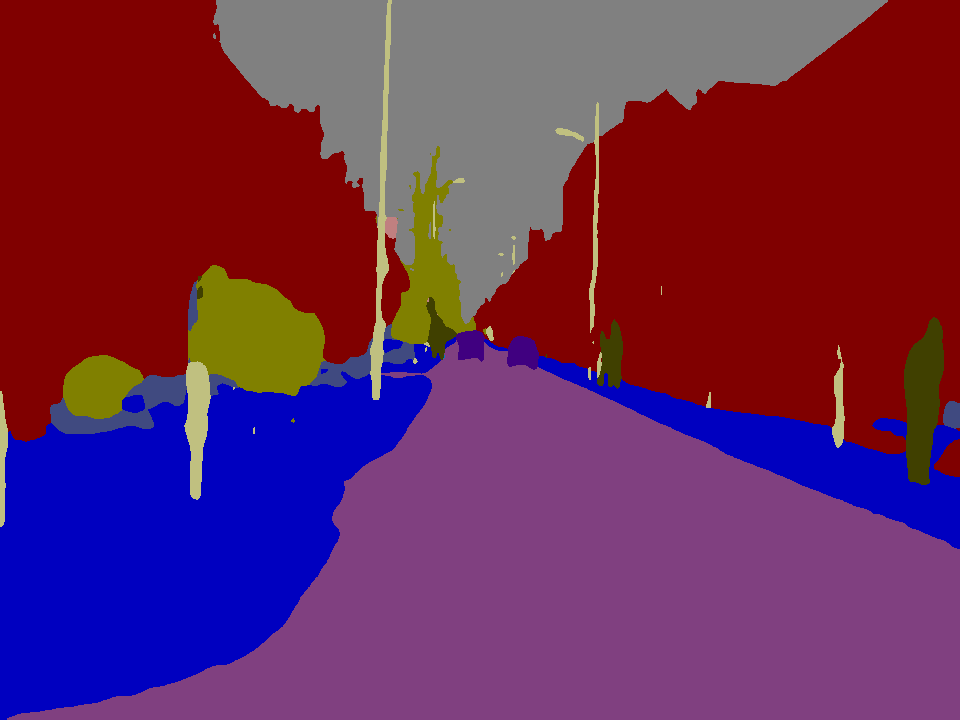}
    \hfill
    \includegraphics[width=\sswt]{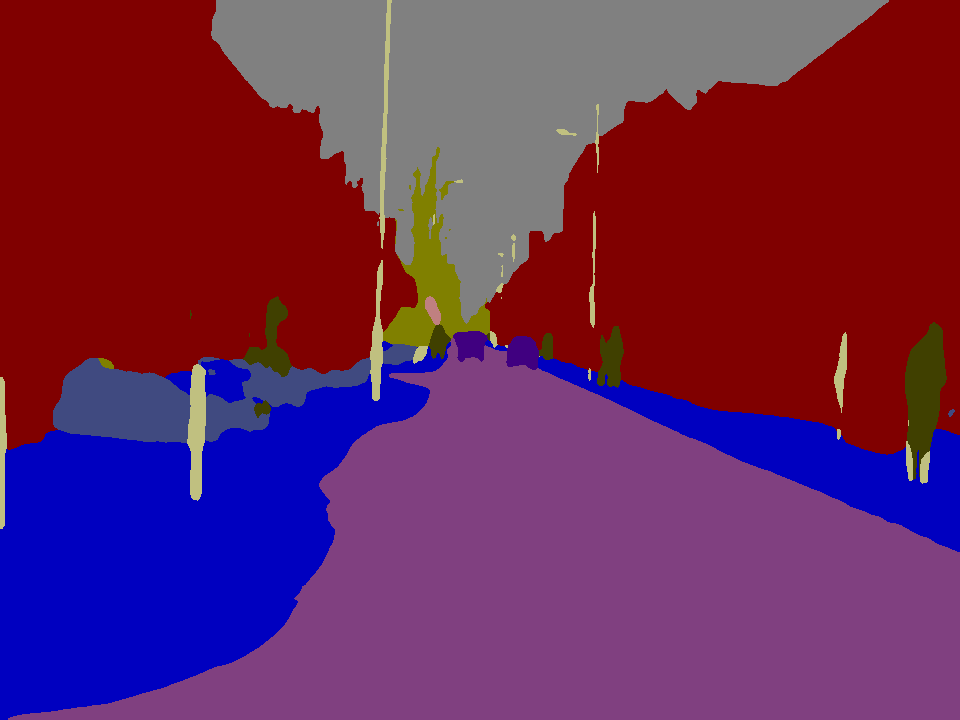}

    \caption{Semantic segmentation results on CamVid test. 
      The columns correspond to input, 
      ground truth, the output of the pyramid model, 
      and the output of the single scale model.
      The most significant improvements
      occur on pixels of classes bus (top) and tree (bottom).
     }
    \label{fig:cv_examples}
\end{figure*}
%------------------------------------------------------------------------
\section{Conclusion}

Real-time performance is a very important trait 
of semantic segmentation models aiming at applications 
in robotics and intelligent transportation systems.
Most previous work in the field involves
custom convolutional encoders trained from scratch, 
and decoders without lateral skip-connections. 
However, we argue that a better 
speed-accuracy trade-off is achieved
with i) compact encoders designed 
for competitive ImageNet performance
and ii) lightweight decoders with
lateral skip-connections.
Additionally, we propose a novel 
interleaved pyramidal fusion scheme
which is able to further improve the results
on large objects close to the camera.
We provide a detailed analysis 
of prediction accuracy and processing time
on Cityscapes and CamVid datasets
for models based on ResNet-18 and MobileNetv2.
Our Cityscapes test submission 
achieves 75.5\% mIoU 
by processing 1024$\times$2048 images 
at 39.9 Hz on a GTX1080Ti. 
To the best of our knowledge, 
this result outperforms 
all previous approaches 
aiming at real-time application.
The source code is available at
\texttt{https://github.com/orsic/swiftnet}.

\section*{Acknowledgment}

This work has been supported by the European Regional
Development Fund under the project System for increased
driving safety in public urban rail traffic (SafeTRAM)
under grant KK.01.2.1.01.0022, and
by European Regional Development Fund
(DATACROSS) under grant KK.01.1.1.01.0009,
and Microblink Ltd.
We would like to thank Josip Krapac for helpful discussions.
The Titan Xp used to train some of the evaluated models
was donated by NVIDIA Corporation.

\pagebreak
{\small
\bibliographystyle{ieee}
\bibliography{orsic19cvpr}

\begin{thebibliography}{10}\itemsep=-1pt

\bibitem{brotsow08prl}
G.~J. Brostow, J.~Fauqueur, and R.~Cipolla.
\newblock Semantic object classes in video: A high-definition ground truth
  database.
\newblock {\em Pattern Recognition Letters}, xx(x):xx--xx, 2008.

\bibitem{chaurasia17vcip}
A.~Chaurasia and E.~Culurciello.
\newblock Linknet: Exploiting encoder representations for efficient semantic
  segmentation.
\newblock In {\em 2017 {IEEE} Visual Communications and Image Processing,
  {VCIP} 2017, St. Petersburg, FL, USA, December 10-13, 2017}, pages 1--4,
  2017.

\bibitem{chen17corr}
L.~Chen, G.~Papandreou, F.~Schroff, and H.~Adam.
\newblock Rethinking atrous convolution for semantic image segmentation.
\newblock {\em CoRR}, abs/1706.05587, 2017.

\bibitem{chen16cvpr}
L.~Chen, Y.~Yang, J.~Wang, W.~Xu, and A.~L. Yuille.
\newblock Attention to scale: Scale-aware semantic image segmentation.
\newblock In {\em 2016 {IEEE} Conference on Computer Vision and Pattern
  Recognition, {CVPR} 2016, Las Vegas, NV, USA, June 27-30, 2016}, pages
  3640--3649, 2016.

\bibitem{cordts15cvpr}
M.~Cordts, M.~Omran, S.~Ramos, T.~Scharw{\"a}chter, M.~Enzweiler, R.~Benenson,
  U.~Franke, S.~Roth, and B.~Schiele.
\newblock The cityscapes dataset.
\newblock In {\em CVPRW}, 2015.

\bibitem{farabet13pami}
C.~Farabet, C.~Couprie, L.~Najman, and Y.~LeCun.
\newblock Learning hierarchical features for scene labeling.
\newblock {\em IEEE transactions on pattern analysis and machine intelligence},
  35(8):1915--1929, 2013.

\bibitem{he15ieee}
K.~He, X.~Zhang, S.~Ren, and J.~Sun.
\newblock Spatial pyramid pooling in deep convolutional networks for visual
  recognition.
\newblock {\em PAMI}, 2015.

\bibitem{he16cvpr}
K.~He, X.~Zhang, S.~Ren, and J.~Sun.
\newblock Deep residual learning for image recognition.
\newblock In {\em CVPR}, pages 770--778, 2016.

\bibitem{he16eccv}
K.~He, X.~Zhang, S.~Ren, and J.~Sun.
\newblock Identity mappings in deep residual networks.
\newblock In {\em ECCV}, pages 630--645, 2016.

\bibitem{howard17corr}
A.~G. Howard, M.~Zhu, B.~Chen, D.~Kalenichenko, W.~Wang, T.~Weyand,
  M.~Andreetto, and H.~Adam.
\newblock Mobilenets: Efficient convolutional neural networks for mobile vision
  applications.
\newblock {\em CoRR}, abs/1704.04861, 2017.

\bibitem{huang17corr}
G.~Huang, S.~Liu, L.~van~der Maaten, and K.~Q. Weinberger.
\newblock Condensenet: An efficient densenet using learned group convolutions.
\newblock {\em arXiv preprint arXiv:1711.09224}, 2017.

\bibitem{huang17cvpr}
G.~Huang, Z.~Liu, L.~van~der Maaten, and K.~Q. Weinberger.
\newblock Densely connected convolutional networks.
\newblock In {\em 2017 {IEEE} Conference on Computer Vision and Pattern
  Recognition, {CVPR} 2017, Honolulu, HI, USA, July 21-26, 2017}, pages
  2261--2269, 2017.

\bibitem{ioffe15icml}
S.~Ioffe and C.~Szegedy.
\newblock Batch normalization: Accelerating deep network training by reducing
  internal covariate shift.
\newblock In {\em ICML}, pages 448--456, 2015.

\bibitem{kingma14corr}
D.~P. Kingma and J.~Ba.
\newblock Adam: {A} method for stochastic optimization.
\newblock {\em CoRR}, abs/1412.6980, 2014.

\bibitem{kreso16gcpr}
I.~Kreso, D.~Causevic, J.~Krapac, and S.~Segvic.
\newblock Convolutional scale invariance for semantic segmentation.
\newblock In {\em GCPR}, pages 64--75, 2016.

\bibitem{kreso17cvrsuad}
I.~Kreso, J.~Krapac, and S.~Segvic.
\newblock Ladder-style densenets for semantic segmentation of large natural
  images.
\newblock In {\em ICCVW CVRSUAD}, pages 238--245, 2017.

\bibitem{lazebnik06cvpr}
S.~Lazebnik, C.~Schmid, and J.~Ponce.
\newblock Beyond bags of features: Spatial pyramid matching for recognizing
  natural scene categories.
\newblock In {\em CVPR}, pages 2169--2178, 2006.

\bibitem{lin17cvpr2}
G.~Lin, A.~Milan, C.~Shen, and I.~D. Reid.
\newblock Refinenet: Multi-path refinement networks for high-resolution
  semantic segmentation.
\newblock In {\em 2017 {IEEE} Conference on Computer Vision and Pattern
  Recognition, {CVPR} 2017, Honolulu, HI, USA, July 21-26, 2017}, pages
  5168--5177, 2017.

\bibitem{lin17cvpr}
T.-Y. Lin, P.~Doll{\'a}r, R.~B. Girshick, K.~He, B.~Hariharan, and S.~J.
  Belongie.
\newblock Feature pyramid networks for object detection.
\newblock In {\em CVPR}, volume~1, page~4, 2017.

\bibitem{long15cvpr}
J.~Long, E.~Shelhamer, and T.~Darrell.
\newblock Fully convolutional networks for semantic segmentation.
\newblock In {\em Proceedings of the IEEE conference on computer vision and
  pattern recognition}, pages 3431--3440, 2015.

\bibitem{loshchilov16arxiv}
I.~Loshchilov and F.~Hutter.
\newblock {SGDR:} stochastic gradient descent with restarts.
\newblock {\em CoRR}, abs/1608.03983, 2016.

\bibitem{luo16nips}
W.~Luo, Y.~Li, R.~Urtasun, and R.~S. Zemel.
\newblock Understanding the effective receptive field in deep convolutional
  neural networks.
\newblock In {\em NIPS}, pages 4898--4906, 2016.

\bibitem{mazzini18bmvc}
D.~Mazzini.
\newblock Guided upsampling network for real-time semantic segmentation.
\newblock In {\em British Machine Vision Conference 2018, {BMVC} 2018,
  Northumbria University, Newcastle, UK, September 3-6, 2018}, page 117, 2018.

\bibitem{mehta18eccv}
S.~Mehta, M.~Rastegari, A.~Caspi, L.~G. Shapiro, and H.~Hajishirzi.
\newblock Espnet: Efficient spatial pyramid of dilated convolutions for
  semantic segmentation.
\newblock In {\em Computer Vision - {ECCV} 2018 - 15th European Conference,
  Munich, Germany, September 8-14, 2018, Proceedings, Part {X}}, pages
  561--580, 2018.

\bibitem{nekrasov18bmvc}
V.~Nekrasov, C.~Shen, and I.~D. Reid.
\newblock Light-weight refinenet for real-time semantic segmentation.
\newblock In {\em British Machine Vision Conference 2018, {BMVC} 2018,
  Northumbria University, Newcastle, UK, September 3-6, 2018}, page 125, 2018.

\bibitem{oquab14cvpr}
M.~Oquab, L.~Bottou, I.~Laptev, and J.~Sivic.
\newblock Learning and transferring mid-level image representations using
  convolutional neural networks.
\newblock In {\em 2014 {IEEE} Conference on Computer Vision and Pattern
  Recognition, {CVPR} 2014, Columbus, OH, USA, June 23-28, 2014}, pages
  1717--1724, 2014.

\bibitem{rasmus15nips}
A.~Rasmus, M.~Berglund, M.~Honkala, H.~Valpola, and T.~Raiko.
\newblock Semi-supervised learning with ladder networks.
\newblock In {\em Advances in Neural Information Processing Systems 28: Annual
  Conference on Neural Information Processing Systems 2015, December 7-12,
  2015, Montreal, Quebec, Canada}, pages 3546--3554, 2015.

\bibitem{romera2018ieee}
E.~Romera, J.~M. Alvarez, L.~M. Bergasa, and R.~Arroyo.
\newblock Erfnet: Efficient residual factorized convnet for real-time semantic
  segmentation.
\newblock {\em IEEE Transactions on Intelligent Transportation Systems},
  19(1):263--272, 2018.

\bibitem{ronneberger15miccai}
O.~Ronneberger, P.~Fischer, and T.~Brox.
\newblock U-net: Convolutional networks for biomedical image segmentation.
\newblock In {\em Medical Image Computing and Computer-Assisted Intervention -
  {MICCAI} 2015 - 18th International Conference Munich, Germany, October 5 - 9,
  2015, Proceedings, Part {III}}, pages 234--241, 2015.

\bibitem{russakovsky15ijcv}
O.~Russakovsky, J.~Deng, H.~Su, J.~Krause, S.~Satheesh, S.~Ma, Z.~Huang,
  A.~Karpathy, A.~Khosla, M.~Bernstein, A.~C. Berg, and L.~Fei-Fei.
\newblock {ImageNet Large Scale Visual Recognition Challenge}.
\newblock {\em International Journal of Computer Vision (IJCV)},
  115(3):211--252, 2015.

\bibitem{sandler18arxiv}
M.~Sandler, A.~G. Howard, M.~Zhu, A.~Zhmoginov, and L.~Chen.
\newblock Inverted residuals and linear bottlenecks: Mobile networks for
  classification, detection and segmentation.
\newblock {\em CoRR}, abs/1801.04381, 2018.

\bibitem{siam18cvpr}
M.~Siam, M.~Gamal, M.~Abdel-Razek, S.~Yogamani, M.~Jagersand, H.~Zhang,
  N.~Vallurupalli, S.~Annamaneni, G.~Varma, C.~Jawahar, et~al.
\newblock A comparative study of real-time semantic segmentation for autonomous
  driving.
\newblock In {\em Proceedings of the IEEE Conference on Computer Vision and
  Pattern Recognition Workshops}, pages 587--597, 2018.

\bibitem{sifre14corr}
L.~Sifre and S.~Mallat.
\newblock Rigid-motion scattering for texture classification.
\newblock {\em CoRR}, abs/1403.1687, 2014.

\bibitem{singh18cvpr}
B.~Singh and L.~S. Davis.
\newblock An analysis of scale invariance in object detection--snip.
\newblock In {\em Proceedings of the IEEE Conference on Computer Vision and
  Pattern Recognition}, pages 3578--3587, 2018.

\bibitem{vallurupalli18cvpr}
N.~Vallurupalli, S.~Annamaneni, G.~Varma, C.~Jawahar, M.~Mathew, and S.~Nagori.
\newblock Efficient semantic segmentation using gradual grouping.
\newblock In {\em The IEEE Conference on Computer Vision and Pattern
  Recognition (CVPR) Workshops}, June 2018.

\bibitem{wang17iccv}
M.~Wang, B.~Liu, and H.~Foroosh.
\newblock Factorized convolutional neural networks.
\newblock In {\em ICCV Workshops}, pages 545--553, 2017.

\bibitem{yu16iclr}
F.~Yu and V.~Koltun.
\newblock Multi-scale context aggregation by dilated convolutions.
\newblock In {\em ICLR}, 2016.

\bibitem{yu17cvpr}
F.~Yu, V.~Koltun, and T.~A. Funkhouser.
\newblock Dilated residual networks.
\newblock In {\em 2017 {IEEE} Conference on Computer Vision and Pattern
  Recognition, {CVPR} 2017, Honolulu, HI, USA, July 21-26, 2017}, pages
  636--644, 2017.

\bibitem{zhang18cvpr}
X.~Zhang, X.~Zhou, M.~Lin, and J.~Sun.
\newblock Shufflenet: An extremely efficient convolutional neural network for
  mobile devices.
\newblock In {\em The IEEE Conference on Computer Vision and Pattern
  Recognition (CVPR)}, June 2018.

\bibitem{zhao2017icnet}
H.~Zhao, X.~Qi, X.~Shen, J.~Shi, and J.~Jia.
\newblock Icnet for real-time semantic segmentation on high-resolution images.
\newblock {\em arXiv preprint arXiv:1704.08545}, 2017.

\bibitem{zhao17iccv}
H.~Zhao, J.~Shi, X.~Qi, X.~Wang, and J.~Jia.
\newblock Pyramid scene parsing network.
\newblock In {\em ICCV}, 2017.

\bibitem{zoph18cvpr}
B.~Zoph, V.~Vasudevan, J.~Shlens, and Q.~V. Le.
\newblock Learning transferable architectures for scalable image recognition.
\newblock In {\em The IEEE Conference on Computer Vision and Pattern
  Recognition (CVPR)}, June 2018.

\end{thebibliography}
}

\end{document}